\documentclass[10pt,twocolumn,letterpaper]{article}

\usepackage[pagenumbers]{cvpr} % To force page numbers, e.g. for an arXiv version

\definecolor{cvprblue}{rgb}{0.21,0.49,0.74}
\usepackage[pagebackref,breaklinks,colorlinks,allcolors=cvprblue]{hyperref}

\usepackage{soul}
\usepackage[utf8]{inputenc}
\usepackage{caption}
\usepackage{amsmath}
\usepackage{amsthm}
\usepackage{booktabs}
\usepackage{algorithm}
\usepackage{algorithmic}
\usepackage[switch]{lineno}

\usepackage[T1]{fontenc}    % use 8-bit T1 fonts
\usepackage{amsfonts}       % blackboard math symbols
\usepackage{nicefrac}       % compact symbols for 1/2, etc.
\usepackage{microtype}      % microtypography

\usepackage{textcomp}
\usepackage{enumitem}
\usepackage{xspace}
\usepackage{subcaption}
\usepackage{multirow}
\usepackage{ifthen}
\usepackage{comment}
\usepackage{rotating}
\usepackage{pifont}
\usepackage[table]{xcolor}

\newcommand{\tool}{\texttt{SentryCam}}

\def\paperID{*****} % *** Enter the Paper ID here
\def\confName{CVPR}
\def\confYear{2026}

\title{Neural Surveillance: Live-Update Visualization of Latent Training Dynamics}

\author{Xianglin Yang\\
National University of Singapore\\
School of Computing\\
{\tt\small xianglin@nus.edu.sg}
\and
Jin Song Dong\\
National Univeristy of Singapore\\
School of Computing\\
{\tt\small dcsdjs@nus.edu.sg}
}

\begin{document}
\maketitle
\begin{abstract}
Monitoring the inner state of deep neural networks is essential for auditing the learning process and enabling timely interventions. 
While conventional metrics like validation loss offer a surface-level view of performance, the evolution of a model's hidden representations provides a deeper, complementary window into its internal dynamics. 
However, the literature lacks a real-time tool to monitor these crucial internal states.
To address this, we introduce \tool, a live-update visualization framework that tracks the progression of hidden representations throughout training. 
\tool{} produces high-fidelity visualizations of the evolving representation space with minimal latency, serving as a powerful dashboard for understanding how a model learns. 
We quantitatively validate the faithfulness of \tool's visualizations across diverse datasets and architectures (ResNet, ViT).
Furthermore, we demonstrate \tool's practical utility for model auditing through a case study on training instability. 
We designed an automated auditing system with geometry-based alerts that successfully identified impending model failure up to 7 epochs earlier than was evident from the validation loss curve. 
\tool's flexible framework is easily adaptable, supporting both the exploratory analysis and proactive auditing essential for robust model development.
\end{abstract}    
\section{Introduction}\label{sec:intro}
Monitoring the inner states of deep neural networks is essential for robust model auditing, enabling convergence analysis and timely debugging~\citep{yang2023deepdebugger}. However, conventional metrics like validation loss are fundamentally \textbf{lagging indicators} of model health. They report on the final outcome of the learning process but offer little insight into the internal mechanics. By the time a problem manifests as a drop in validation performance, the underlying cause—such as a collapse of the feature space—may already be irreversible.

To achieve a \emph{proactive} and more \emph{granular} understanding, we must inspect the evolution of the model's hidden representations directly. Analyzing these internal dynamics provides two critical advantages. First, it serves as a \textbf{leading indicator} of model failure. Pathological geometric events often precede any noticeable drop in accuracy, offering a crucial window for intervention. Second, it provides a granular view into the learning process itself, allowing practitioners to audit how a model adapts to data~\citep{chen2023real}, validate the effectiveness of new techniques, or inspire novel approaches~\citep{parvaneh2022active}. By complementing reactive metrics with this direct, proactive view into the latent space, developers can achieve a more holistic audit, leading to faster and more reliable development cycles.

However, analyzing high-dimensional, evolving representations in a manner that is both interpretable and computationally efficient poses a significant challenge. While Trace Visualization via dimensionality reduction has emerged as a powerful paradigm for this problem~\citep{rauber2016visualizing,yang2022deepvisualinsight,yang2022temporality}, its transition from a post-hoc analysis technique to a viable, real-time auditing tool imposes a strict set of practical requirements. Success is not merely about achieving high visualization fidelity—i.e., preserving spatial and temporal data structures. For a system to be truly useful for live auditing, we argue it must also embody three crucial \textbf{practicality desiderata}: it must be \textbf{(1) Automatic}, operating without tedious per-dataset tuning; \textbf{(2) Live-Update}, providing immediate feedback for timely intervention; and \textbf{(3) Extensible}, adapting to dynamic scenarios like continual learning.

It is precisely these practical requirements that expose the limitations of prior parametric dimension reduction works. Many systems demand cumbersome manual hyperparameter tuning (e.g., TimeVis~\citep{yang2022temporality}), while others are architecturally incapable of providing the live updates essential for real-time monitoring (e.g., DVI~\citep{yang2022deepvisualinsight}, M-PHATE~\citep{gigante2019visualizing}). Without a framework that embodies these principles, the current tools are either too cumbersome or too slow for proactive auditing, forcing them to risk significant computational resources on training runs that are destined to fail.

To address this gap, we introduce \tool{}, a visual analysis framework designed for the real-time monitoring of neural network training. 
While \tool{} builds on the paradigm of parametric dimensionality reduction~\citep{sainburg2021parametric,yang2022deepvisualinsight}, its core novelty lies in overcoming the two critical challenges of applying this technique in a live, streaming context: (1) efficient spatio-temporal graph construction and (2) stable projection learning.
Our framework solves these challenges with a set of integrated solutions. For graph construction, a \textit{selective working memory} manages temporal complexity, a principled \textit{optimal sampling ratio search} ensures real-time performance without sacrificing fidelity. 
For stable projection, we leverage a novel \textit{hybrid normalization scheme} that makes our projection model immune to the severe distributional shifts inherent in the data stream.

We rigorously validate \tool{} across a wide range of architectures (CNNs, Transformers) and datasets. The results demonstrate that \tool{} produces visualizations of a quality comparable to or superior to state-of-the-art baselines while being significantly more efficient and fully automated.
Furthermore, we demonstrate \tool’s practical utility for model auditing through a case study. We designed an automated auditing system with geometry-based alerts that successfully identified impending model failure up to 7 epochs earlier than was evident from the validation loss curve.
\tool's flexible framework is easily adaptable, supporting both the exploratory analysis and proactive auditing essential for robust model development.

In summary, our contributions are three-fold:
\begin{itemize}[leftmargin=*]
\item \textbf{A Set of Principles for Practical Visual Monitoring:} We define and justify three core requirements for real-time DNN analysis systems: \emph{automation}, \emph{live updates}, and \emph{extensibility}.
\item \textbf{A Novel Visualization Framework, \tool{}:} We propose a framework that embodies these principles by overcoming the critical challenges of dynamic graph construction and stable projection learning through innovations in temporal memory management, principled optimal sampling, and adaptive normalization.
\item \textbf{A Demonstration of Practical Utility:} We provide extensive quantitative validation and present a case study showing \tool{}'s effectiveness in providing early, actionable warnings for proactive model auditing.
\end{itemize}

\section{The Challenge of Live Latent Visualization}\label{sec:challenges}
\tool{} is built upon the paradigm of Parametric Dimension Reduction (PDR), where a parametric model is trained to project high-dimensional data into an intuitive low-dimensional space. While powerful for post-hoc analysis, adapting this pipeline into a real-time auditing tool is a non-trivial task. Here, we outline the standard PDR pipeline and the critical challenges that arise when attempting to apply it in a live, streaming context. (See Appendix~\ref{app:background} for the detailed parametric training process.)

\noindent\textbf{Phase \ding{182}: Representation Extraction.} The hidden activations are captured at each training checkpoint. 

\noindent\textbf{Phase \ding{183}: Spatio-Temporal Graph Construction.} In a live setting, the goal is to build a graph that combines the current epoch's spatial structure with its temporal relation to the past. Attempting this results in a major bottleneck: \textbf{Computational Scalability}. To provide both the current and historical contexts, the graph must incorporate nodes from all epochs. Naively including all data causes the graph to grow linearly, making its construction and processing computationally intractable for real-time updates.

\noindent\textbf{Phase \ding{184}: Projection Learning.} A parametric model (e.g., an autoencoder) is trained to project the complex graph into an intuitive 2D space. Here, the primary challenge is \textbf{Training Instability}. Our input is inherently heterogeneous, comprising nodes from different training epochs, each with a unique statistical distribution. This heterogeneity causes standard normalization techniques (like Batch Normalization~\citep{ioffe2015batchnormalizationacceleratingdeep}) to compute noisy, unstable statistics, leading to poor convergence and a low-quality final visualization.

These challenges explain why prior visualization methods have been largely confined to post-hoc analysis. Addressing them directly is the primary motivation for our work and the key to enabling the truly live-update auditing that previous systems could not achieve.
\section{Method}\label{sec:method}
\begin{figure*}[t]
    \centering
    \includegraphics[width=0.80\textwidth]{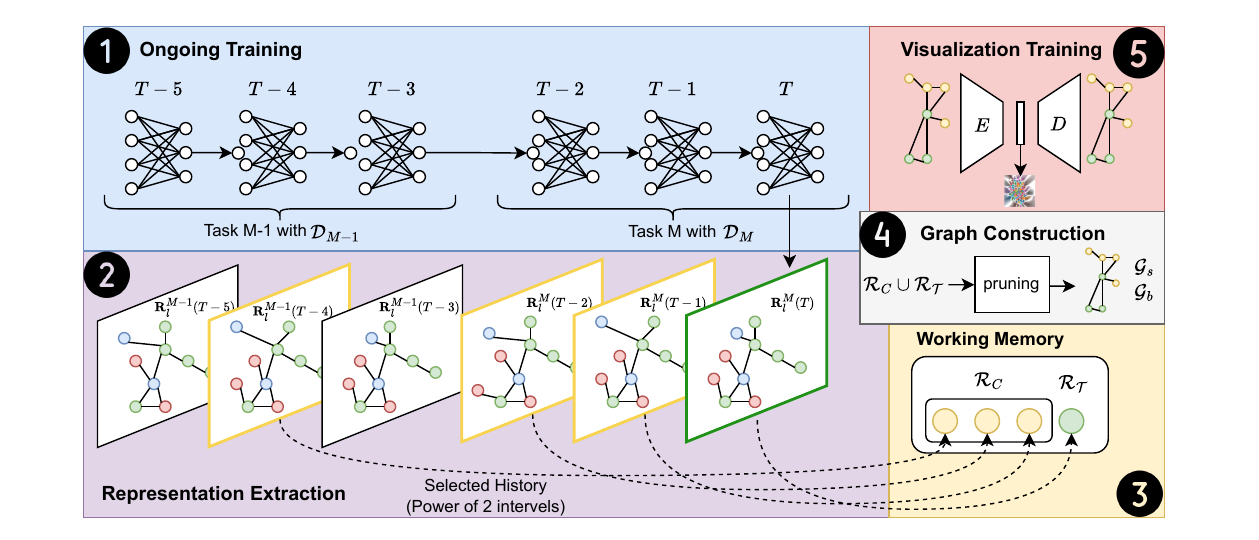}
    \caption{\textbf{Overview of \tool}. First, as the subject model trains (stage \ding{182}), our framework intercepts its hidden representations at each checkpoint (stage \ding{183}). 
Next, we assemble a working memory of nodes, comprising the current representations and a curated set of historical ones. 
From this memory, we construct a composite graph that encodes both spatial and temporal relationships (stage \ding{184}). 
To ensure real-time performance, this graph is immediately downsampled using our novel density-guided algorithm (stage \ding{185}). 
Finally, a specialized autoencoder projects the resulting graph into a low-dimensional space to produce the final interpretable visualization (stage \ding{186}).}
    \label{fig:overview}
\end{figure*} 

% \tool{} generates a live visualization of a model's evolving representation space through the real-time, multi-stage pipeline shown in Figure~\ref{fig:overview}. In this section, we explain how we address those challenges in Section~\ref{sec:challenges}.
\tool{} generates a live visualization of a model's evolving representation space through the real-time, multi-stage pipeline shown in Figure~\ref{fig:overview}. This pipeline's real-time capability is achieved through a set of core innovations designed to solve the challenges outlined in Section~\ref{sec:challenges}. We detail these solutions in the following sections.

\subsection{Notations}
For the model to be audit,
we consider a general training setting where a neural network $f_\theta: \mathbb{R}^s \rightarrow \mathbb{R}^C$, referred to as the subject model, is trained progressively on a sequence of $M$ tasks. 
The training data consists of a collection of datasets $\mathcal{D} = \{\mathcal{D}_1, \mathcal{D}_2, \dots, \mathcal{D}_M\}$, where each task $m$ has an associated dataset $\mathcal{D}_m = \{(x^m_i, y^m_i)\}_{i=0}^n$.
When $M=1$, this reduces to standard supervised learning. 
For $M>1$, it corresponds to a continual learning setting.
This make sure our framework is extensible for visualization.

\noindent\textbf{Stage \ding{182}: Representation Extraction.} Given a model with $L$ layers, we can dissect it at any layer $l$ into two segments: a feature extractor $f_{1:l}:\mathbb{R}^s\rightarrow\mathbb{R}^r$ and a classification head $f_{l:L}:\mathbb{R}^r\rightarrow\mathbb{R}^C$.
We denote the set of hidden representations (activations) from layer $l$ for dataset $\mathcal{D}_m$ at the training epoch $t$ as $\mathbf{R}^m_l(t)=f_{1:l}(\mathcal{D}_m)$.
% with shape of (n, r), n is the number of data and r is the embedding dimension.

\subsection{Dynamic Graph Construction}\label{sec:graph_cons}
\paragraph{Stage \ding{183}-\ding{184}: Working Memory Construction.}
To construct a graph that captures temporal evolution, we must have a set of nodes from the past epochs. A naive approach of including all historical representations is computationally infeasible, as the number of nodes would grow linearly with training time, $\mathcal{O}(t)$. To overcome this, we introduce a \textbf{selective working memory} mechanism that efficiently curates a set of historical nodes, $\mathbf{R}_\mathcal{C}$.

Our approach uses a deterministic, \textbf{Logarithmic Time Aggregation} strategy. At any given epoch $t$, we aggregate representation sets from past epochs $t'$ that fall on power-of-two intervals from the present. The set of historical nodes is thus defined as:$
    \mathbf{R}_\mathcal{C} = 
    \bigcup_{t-t'=2^n, n\in\mathbb{N}^+, t' \in [t-1]} \mathbf{R}_l(t')
$.
This strategy elegantly balances the preservation of short- and long-term memory. As $t$ increases, the intervals between selected past epochs ($t'$) expand, allowing the memory to retain context from the distant past while still densely including recent history. This reduces the number of historical time steps from $\mathcal{O}(t)$ to $\mathcal{O}(\log t)$, a critical optimization for real-time performance. The final working memory is composed of these historical nodes along with the current representations, $\mathbf{R}_\mathcal{T} = \mathbf{R}_l(t)$, forming the complete vertex set for our dynamic graph.

\begin{figure}[h!]
     \centering
     \begin{subfigure}[b]{0.23\linewidth}
         \centering
         \includegraphics[width=\textwidth]{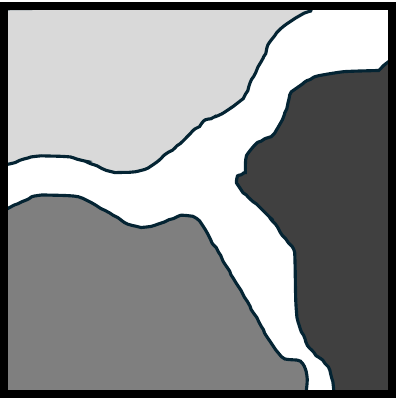}
         \caption{}
         \label{fig:demo-db}
     \end{subfigure}
     % \hfill
     \begin{subfigure}[b]{0.23\linewidth}
         \centering
         \includegraphics[width=\textwidth]{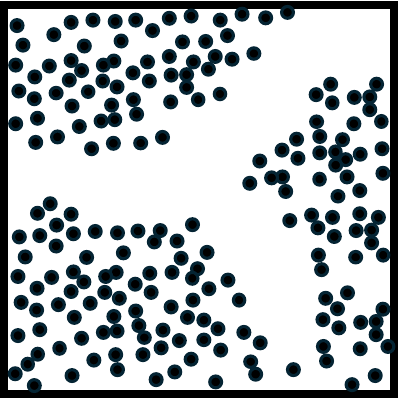}
         \caption{}
         \label{fig:demo-all-data}
     \end{subfigure}
     % \hfill
     \begin{subfigure}[b]{0.23\linewidth}
         \centering
         \includegraphics[width=\textwidth]{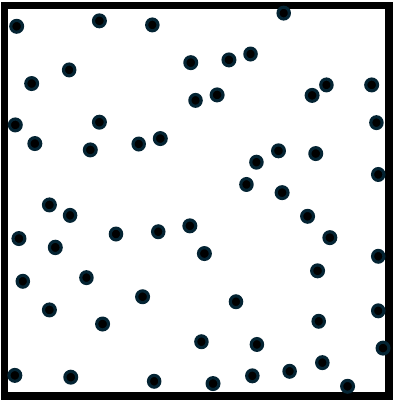}
         \caption{}
         \label{fig:demo-downsampling}
     \end{subfigure}
     \caption{\textbf{The role of sample density in revealing underlying data structure.} (a) The ground truth distribution of three distinct classes. (b) A sufficiently dense sample allows for the faithful reconstruction of the three-cluster structure. (c) A sparse sample provides insufficient information, causing the underlying topology to be lost.}
    \label{fig:demo}
\end{figure}

\noindent\textbf{Stage \ding{185} (pruning): Optimal Sampling Rate Search.}\label{sec:sampling}
While our working memory manages temporal complexity, the number of data points per time step remains a critical performance bottleneck. The challenge is to find what is the minimal subset of points via \emph{random sampling} that still faithfully captures the data's underlying topology (as demonstrated in Figure~\ref{fig:demo}). 

\noindent\textbf{Hypothesis and Empirical Validation.}
We hypothesize that the loss of topological structure with sampling is not gradual and linear, but instead exhibits a sharp, non-linear collapse at a critical density threshold—a phenomenon we term the \emph{Topological Tipping Point}. As illustrated in Figure~\ref{fig:demo}(c), once this tipping point is crossed, essential geometric information is irrecoverably ``shattered'', leading to a catastrophic collapse in visualization quality.

\begin{figure}[h!]
     \centering
     \begin{subfigure}[b]{0.4\linewidth}
         \centering
         \includegraphics[width=\textwidth]{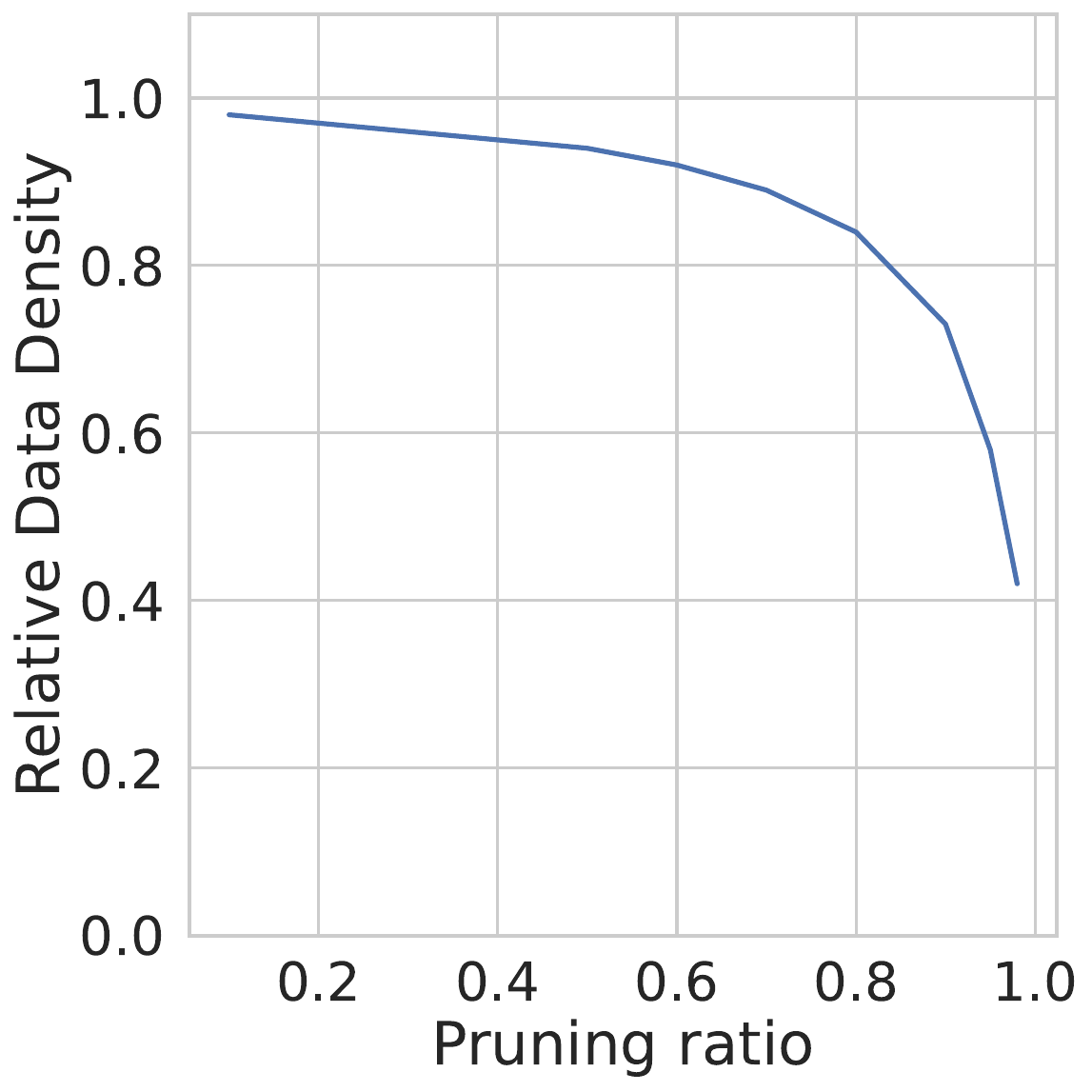}
         \caption{}
         \label{fig:sentrycam-pruning-density}
     \end{subfigure}
     \begin{subfigure}[b]{0.5\linewidth}
         \centering
         \includegraphics[width=\textwidth]{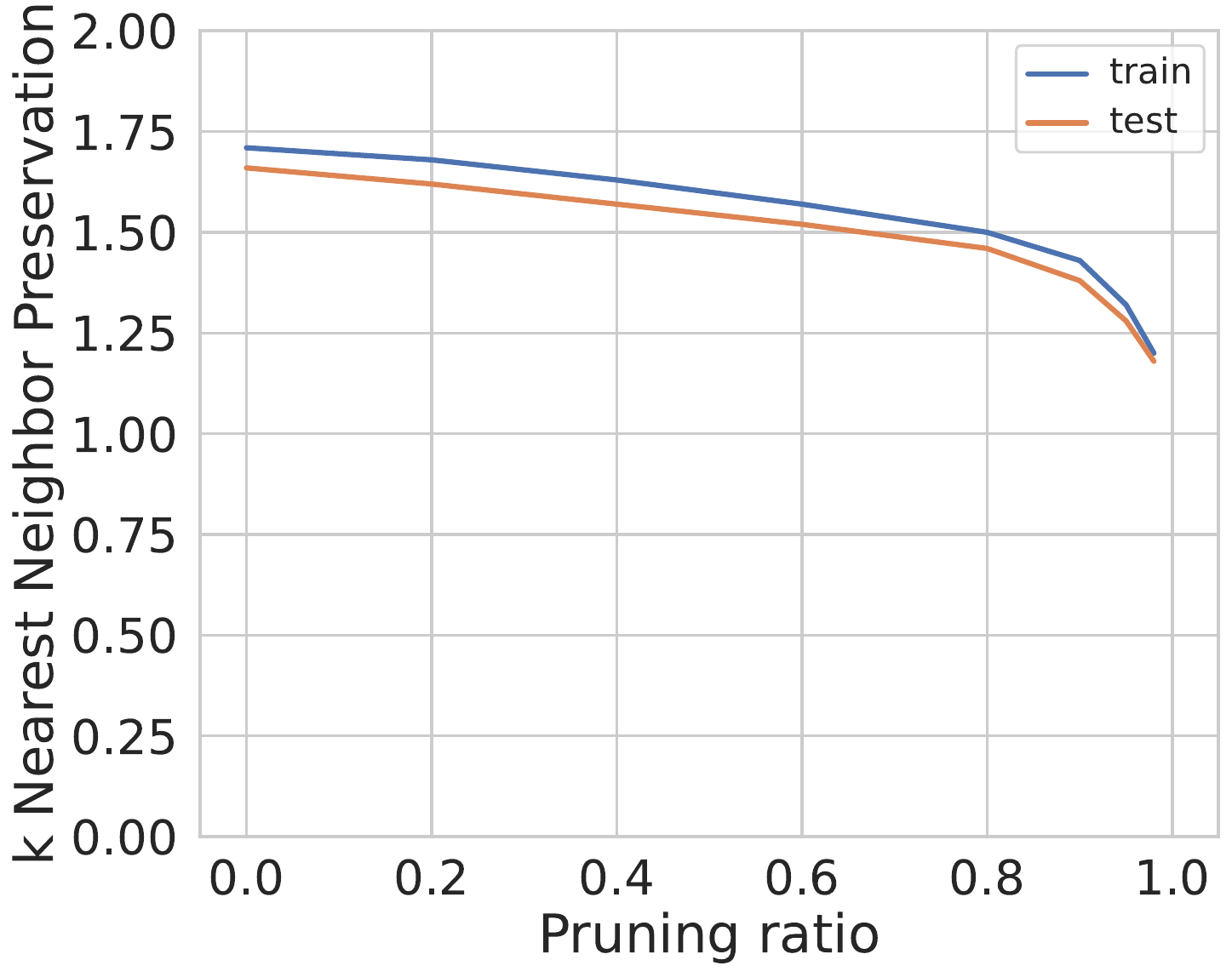}
         \caption{}
         \label{fig:sentrycam-pruning-nn}
     \end{subfigure}
     \caption{(a) The relationship between data sampling ratio and relative data density. (b) The relationship between the sampling ratio and the k nearest neighbor preservation (k=15) of the resulting low-dimensional embedding.}
    \label{fig:sentrycam-empirical-observation}
\end{figure}

To test this hypothesis, we conducted an empirical study, progressively downsampling a dataset while measuring both visualization quality and our proposed proxy metric: relative data density\footnote{Defined as the ratio of average k-NN distances in the sampled versus the original set.}. The results in Figure~\ref{fig:sentrycam-empirical-observation} provide clear evidence for our hypothesis. The visualization quality (b) exhibits the exact ``plateau and cliff'' structure we predicted: a wide region of stability followed by a sudden drop. More importantly, the curve for our fast proxy metric (a) perfectly mirrors this behavior, confirming that it is a highly reliable indicator for detecting the edge of the stability plateau.

This empirical result is not an isolated artifact but is grounded in formal theory. As we prove in Section~\ref{sec:theory}, this phase transition is a predictable property of manifold sampling (Proposition~\ref{prop:phase}), and our relative density metric is directly related to the sufficient conditions for topology preservation (Corollary~\ref{cor:relden}).

\noindent\textbf{Solution: An Automated Search Algorithm.}
Based on this strong empirical and theoretical foundation, we designed a novel algorithm to automatically find the optimal random sampling rate. The algorithm acts as an efficient ``tipping point detector,'' using the validated proxy metric (relative data density) in a binary search to quickly identify the most aggressive sampling ratio that remains safely on the stability plateau. This automates the creation of a computationally efficient data representation—obtained via random sampling at the identified optimal rate—that is still guaranteed to preserve its essential topological structure. This capability is critical for any robust, real-time monitoring system, and the detailed algorithm is described in Appendix~\ref{sec:detailed-density-guided-sampling}.

\noindent\textbf{Stage \ding{185}: Graph Construction.}
Once the working memory of nodes ($\mathbf{R}_\mathcal{T} \cup \mathbf{R}_\mathcal{C}$) has been assembled and pruned, we construct a composite graph $\mathcal{G} = \mathcal{G}_s \cup \mathcal{G}_b$ by defining two distinct sets of weighted edges to capture both spatial and temporal relationships.

\textit{Spatial Edges ($\mathcal{G}_s$).} To preserve the topology of the current representations ($\mathbf{R}_\mathcal{T}$), we construct a $k$-nearest neighbor graph following the UMAP~\citep{gigante2019visualizing}. 
An edge $(r_i, r_j, p(r_i, r_j))$ is created between each representation $r_i$ and its $k$-nearest neighbors $r_j \in \mathcal{N}^k(r_i)$. 
The edge weight $p(r_i, r_j)$ is a fuzzy similarity score derived from the distance between the points, as specified in Eq.~\ref{eq:pij} in Appendix~\ref{app:background}. This graph, $\mathcal{G}_s$, encodes the intraslice structure of the current time step.

\textit{Temporal Edges ($\mathcal{G}_b$).} To capture the model's evolution, we form a bipartite graph $\mathcal{G}_b$ connecting the current representations $\mathbf{R}_\mathcal{T}$ to the historical representations $\mathbf{R}_\mathcal{C}$. 
A critical challenge here is that representation vectors from different epochs are not directly comparable using standard distance metrics due to shifts in their statistical distributions. To create meaningful and controllable temporal edges, we use \textbf{cosine similarity} as the weight function: $\mathcal{G}_b= \{(r, r', \cos(r, r')) \mid r \in \mathbf{R}_\mathcal{T}, r' \in \mathbf{R}_\mathcal{C}\}$.
We chose cosine similarity because it emphasizes directional change over magnitude, providing a more robust and normalized measure of how representation vectors evolve over time. This graph, $\mathcal{G}_b$, encodes the interslice relationships that are crucial for visualizing dynamics.

\subsection{Stable Projection Learning}
\noindent\textbf{Stage \ding{186}:Visualization Generation.}
Once the composite graph $\mathcal{G} = \mathcal{G}_s \cup \mathcal{G}_b$ is constructed, we learn a low-dimensional projection using an autoencoder~\cite{hinton2006reducing}. 
The autoencoder is optimized with a hybrid loss function that combines the UMAP cost (Eq.~\ref{eq:umap} in Appendix~\ref{app:background}) to preserve topological structure with a Mean Squared Error term for accurate feature reconstruction.

A critical challenge is \textbf{training instability} caused by the heterogeneity of input, which causes standard normalization techniques (like Batch Normalization~\citep{ioffe2015batchnormalizationacceleratingdeep}) to compute noisy, unstable statistics, leading to poor convergence and a low-quality final visualization.
To address this, we introduce a novel \textbf{hybrid normalization scheme} that couples Group Normalization (GN)~\cite{wu2018groupnormalization} with BN. The key insight is that GN operates at the instance level, making it agnostic to batch statistics. Our architecture strategically places a GN layer \textit{before} the BN layers in the autoencoder. For instance, the first layer of the encoder is defined as: $z_1 = \text{ReLU}(\text{BN}(\text{GN}(W_1 x_1 + b_1)))$.
By first stabilizing each instance with GN, we mitigate the severe distributional shifts before the batch is processed by BN. This synergistic coupling allows our projection model to retain the fast convergence benefits of BN while overcoming its vulnerability to heterogeneous data. The result is a highly stable and reliable training process, which is essential for a real-time visualization system.
% ----- Packages -----
% \usepackage{bm}

% ----- Theorem Environments -----
\newtheorem{theorem}{Theorem}
\newtheorem{lemma}[theorem]{Lemma}
\newtheorem{corollary}[theorem]{Corollary}
\theoremstyle{definition}
\newtheorem{definition}[theorem]{Definition}
\newtheorem{proposition}[theorem]{Proposition}

% ----- Macros -----
\newcommand{\R}{\mathbb{R}}
\newcommand{\dist}{\operatorname{dist}}
\newcommand{\Dg}{\mathrm{Dg}}
\newcommand{\Rips}{\mathrm{Rips}}
\newcommand{\Cech}{\check{C}}
\newcommand{\op}{\mathop{}}
\newcommand{\E}{\mathbb{E}}
\newcommand{\Prob}{\mathbb{P}}
\newcommand{\dd}{\mathrm{d}}
\newcommand{\diag}{\operatorname{diag}}
\newcommand{\bott}{d_B}
\newcommand{\Wtwo}{W_2}
\newcommand{\vol}{\operatorname{vol}}

\section{Topology Preservation Under Sampling}
\label{sec:theory}
We formalize when a downsampled set of latent points preserves the intrinsic topology of the representation manifold and explain the empirically observed \emph{topology tipping point} as identified in Section~\ref{sec:empirical-ob}. 
Our guarantees convert a simple, measurable quantity---the average relative $k$-NN distance---into a sufficient condition for homotopy preservation.

\noindent\textbf{Setup.}
Let $\mathcal{M}\subset\mathbb{R}^h$ be a compact $\mathcal{C}^2$ $d$-manifold with reach $\tau$\footnote{The reach measures the regularity of the manifold, which is usually used for manifold estimation~\citep{aamari2019estimatingreachmanifold}.}$>0$.
At a fixed slice, we have embeddings $X=\{x_i\}_{i=1}^N\subset\mathcal{M}$ and a downsample $S\subset X$ of size $n=\lfloor rN\rfloor$.
Write the covering radius $\varepsilon(S)=\sup_{y\in\mathcal{M}}\min_{s\in S}\|y-s\|$ and the average $k$-NN distance
$\overline{d}_k(A)=|A|^{-1}\!\sum_{a\in A}\|a-\mathrm{NN}_k(a;A)\|$.
Define the scale-free density proxy
\begin{equation}
\mathrm{RelDen}_k(S;X)\ :=\ \frac{\overline{d}_k(S)}{\overline{d}_k(X)}.
\label{eq:relden_short}
\end{equation}

\noindent\textbf{Goal.}
Give a \emph{practical} condition (in terms of \eqref{eq:relden_short}) guaranteeing that a proximity complex on $S$
(\v{C}ech/Rips/UMAP) preserves the homotopy type of $\mathcal{M}$ with high probability.

\subsection{Guarantee of Topology Preservation via Relative Density Metric}
\noindent\textbf{Reach-based topological guarantee}
\label{subsec:reach-guarantee}
The following classical result ties topological correctness to sampling density relative to the feature size (reach).

\begin{theorem}[Homotopy preservation under $\varepsilon$-density~\citep{Hatcher2002,10.5555/3116660.3117013}]
\label{thm:reach}
There exists a universal constant $c\in(0,1)$ such that if $d_H(S,\mathcal{M})=\varepsilon(S)<c\,\tau$, then there is a non-empty interval of radii $\alpha$ for which the \v{C}ech complex $\check{C}_\alpha(S)$ is homotopy equivalent to $\mathcal{M}$. Moreover, the Vietoris--Rips complex $R_\rho(S)$ is also homotopy equivalent to $\mathcal{M}$ for $\rho$ in an interval determined by standard interleavings.
\end{theorem}
Theorem~\ref{thm:reach} reduces topology preservation to keeping the covering radius $\varepsilon(S)$ below a constant fraction of the reach $\tau$. The proofs is in Appendix~\ref{app:reach-proof}.

\noindent\textbf{From $k$-NN distance to covering radius}
\label{subsec:knn-to-eps}
While $\varepsilon(S)$ and $\tau$ are not directly observable, the average $k$-NN distance is, and it controls $\varepsilon(S)$ up to constants under mild regularity.

\begin{lemma}[Equivalence up to constants]
\label{lem:eps-knn}
Assume the points are sampled i.i.d.\ from a density on $\mathcal{M}$ that is bounded and bounded away from zero on $\mathcal{M}$.
Fix $k\ge 1$. There exist constants $C_1,C_2>0$ depending only on $(\mathcal{M},\tau,d)$ and the density bounds such that, with high probability,
\[
C_1\,\overline{d}_k(S)\ \le\ \varepsilon(S)\ \le\ C_2\,\overline{d}_k(S).
\]
\end{lemma}
Intuitively, $\overline{d}_k(S)\asymp (k/n)^{1/d}$ and $\varepsilon(S)\asymp (\log n/n)^{1/d}$~\citep{penrose2003random}. The proof of Lemma~\ref{lem:eps-knn} is in Appendix~\ref{app:eps-knn-proof}.

\noindent\textbf{Operational criterion via a relative density threshold.}
Combining Theorem~\ref{thm:reach} and Lemma~\ref{lem:eps-knn} yields a practical, scale-free test.

\begin{corollary}[Relative-density sufficient condition]
\label{cor:relden}
Let $\eta^\star$ be any constant such that
\begin{equation}
(C_2/C_1)\,\eta^\star\,\overline{d}_k(X)\ <\ c\,\tau.
\label{eq:eta-star}
\end{equation}
If $\mathrm{RelDen}_k(S;X)\le \eta^\star$, then there exists a radius for which a proximity complex on $S$ is homotopy equivalent to $\mathcal{M}$ with high probability.
\end{corollary}
Because $\tau$ is unknown, we select a conservative \emph{working} threshold $\eta_{\mathrm{th}}<\eta^\star$ (e.g., $\eta_{\mathrm{th}}=0.8$ in our experiments), and accept a downsample if $\mathrm{RelDen}_k(S;X)\le \eta_{\mathrm{th}}$.
This condition is easy to verify and empirically aligns with the stability plateau observed in our figures.

\subsection{The Topology Tipping Point }
\label{subsec:tipping}
As the sampling ratio $r$ decreases, $n=rN$ shrinks and $\overline{d}_k(S)$ increases monotonically, so $\mathrm{RelDen}_k(S;X)$ increases stochastically.

\begin{proposition}[Phase transition \citep{cohen2005stability,chazal2016structure}]
\label{prop:phase}
There exists a critical ratio $r_\mathrm{tip}\in(0,1)$ such that for $r>r_\mathrm{tip}$ the sufficient condition in Cor.~\ref{cor:relden} holds with high probability (stable topology), while for $r<r_\mathrm{tip}$ it fails (topology changes). Consequently, $k$-NN neighborhood preservation and Betti numbers remain near-constant for $r>r_\mathrm{tip}$ and degrade sharply once $r$ crosses $r_\mathrm{tip}$ from above.
\end{proposition}

Proposition~\ref{prop:phase} explains the \emph{tipping point} behavior: a wide pre-tipping plateau followed by an abrupt topological collapse once the covering radius breaches a fixed fraction of the reach (proof in App.~\ref{app:phase-proof}).

% 1. what evaluation?
% 2. why?
% 3. exp figure explain
% 4. result explain(what phenominon) 
% 5. explain the reason
\section{Experiments}\label{sec:exp}
In this section, we conduct a series of experiments to answer three key research questions:
\begin{itemize}[leftmargin=*]
\item \textbf{RQ1 (Efficiency):} How efficient is \tool{} in generating live-update visualizations compared to baseline methods?
\item \textbf{RQ2 (Visual Quality):} To what extent does \tool{} maintain state-of-the-art visualization quality while achieving its goals of real-time efficiency and extensibility?
\item \textbf{RQ3 (Utility):} Can \tool's visualizations provide practical, timely alerts for auditing the training process?
\end{itemize}

\subsection{Experimental Setup}
To test the generalization ability of \tool, we evaluate it across a diverse range of training scenarios.

\noindent\textbf{Datasets.} We use three standard image classification datasets with varying complexity, image sizes, and class numbers: CIFAR-10~\cite{Krizhevsky09learningmultiple}, CIFAR-100~\cite{Krizhevsky09learningmultiple}, and Food-101~\cite{bossard2014food}. Further details are in Appendix~\ref{sec:datasets}.

\noindent\textbf{Subject Models for Visualization}
Our framework is model-agnostic, operating on the latent representations of \emph{any} neural network. For clear and rigorous validation, we demonstrate its utility on the foundational task of image classification using two common architectural families: CNN-based models (e.g., ResNet variants) and Transformer-based models (e.g., ViT variants). Noted that the approach is not limited to classifiers; it can be readily applied to analyze the internal states of generative models (e.g., tracking latents in a diffusion model's U-Net~\citep{ho2020denoisingdiffusionprobabilisticmodels}) or large foundation models (e.g., auditing embeddings in SAM~\citep{kirillov2023segment}). 
% This task provides an unambiguous benchmark for evaluating the quality and fidelity of the visualization. 

To evaluate performance under diverse conditions, we use two settings: (1) training from scratch (ResNet-18/34, small ViT on CIFAR-10/100) and (2) fine-tuning pre-trained models (ResNet-50, ViT/B-16 on FOOD-101). We extract representations from the penultimate layer. All experiments were conducted on one NVIDIA A800 GPU. Details are in Appendix~\ref{sec:subject-models}.

\noindent\textbf{Training Configuration for Visualization Models}  
The visualization model is an autoencoder that transforms high-dimensional data into compact representations while preserving its structure. 
The encoder reduces dimensions by fully connected layers: $(d, \lfloor d/2 \rfloor, \lfloor d/4 \rfloor, \lfloor d/8 \rfloor, \lfloor d/16 \rfloor, 2)$, and the decoder mirrors this structure for reconstruction. 
The model is trained using the Adam optimizer (learning rate $1 \times 10^{-2}$, weight decay $1 \times 10^{-5}$) with a StepLR scheduler (step size 4, gamma 0.1) to balance convergence speed and stability. 
This configuration ensures robust performance across scenarios.

\noindent\textbf{Baselines.} We compare \tool{} against two state-of-the-art baselines: (1) DVI~\citep{yang2022deepvisualinsight}, a parametric method using sequential training, and (2) TimeVis \citep{yang2022temporality}, a post-hoc, autoencoder-based approach. We follow the experimental setups described in their original papers.

\subsection{RQ1: Efficiency}

\begin{table*}[t!]
\centering
\resizebox{\linewidth}{!}{
\begin{tabular}{c|c|ccccccccc}
\toprule
\multirow{2}{*}{\textbf{Model}} &
  \multirow{2}{*}{\textbf{Method}} &
  \multicolumn{3}{c}{\textbf{CIFAR-10}} &
  \multicolumn{3}{c}{\textbf{CIFAR100}} &
  \multicolumn{3}{c}{\textbf{FOOD-101}} \\
\cmidrule(lr){3-5} \cmidrule(lr){6-8} \cmidrule(lr){9-11} 
 &
   &
  AVT/ATT &
  Max Delay/ATT &
  AVG Delay/ATT &
  AVT/ATT &
  Max Delay/ATT &
  AVG Delay/ATT &
  AVT/ATT &
  Max Delay/ATT &
  AVG Delay/ATT \\
\midrule
\multirow{3}{*}{CNN based} & DVI       & 10.6 & 2120.7 & 1165.2 & 7.1 & 1619.4 & 913.2 & 1.8 & 22.1 & 21.8 \\
                           & TimeVis   & -    & 478.6  & 478.7  & -   & 447.6  & 447.6 & -   & 37.4 & 37.4 \\
\rowcolor[gray]{0.95} \cellcolor{white}& \tool{} & 6.2  & 8.2    & 6.2    & 3.8 & 5.3    & 3.8   & 1.0 & 1.0  & 1.0  \\
\midrule
\multirow{3}{*}{ViT based} & DVI       & 4.3  & 865.4  & 534.5  & 2.4 & 472.4  & 337.5 & 1.0 & 21.1 & 21.0 \\
                           & TimeVis   & -    & 297.3  & 297.3  & -   & 143.5  & 143.5 & -   & 13.0 & 13.0 \\
\rowcolor[gray]{0.95} \cellcolor{white}& \tool{} & 2.4  & 2.7    & 2.4    & 1.3 & 1.4    & 1.3   & 0.7 & 0.7  & 0.7 \\
\bottomrule
\end{tabular}
}
\caption{\textbf{Efficiency Comparison of Visualization Methods on CIFAR10, CIFAR100, and FOOD101 Datasets.} All values are normalized by the Average Training Time per Epoch (ATT).}
\label{tb:efficiency}
\end{table*}

\noindent\textbf{Evaluation Setup.}
To assess \tool's suitability for live monitoring, we evaluate its computational efficiency. We move beyond hardware-dependent raw timings and instead report normalized metrics that measure practical feasibility. All metrics are presented as a ratio relative to the Average Training Time per Epoch (ATT) of the subject model, where a value less than 1.0 indicates that a task finishes before the next training epoch completes. We report\footnote{We report the raw clock wall time in the Appendix~\ref{app:wall-clock}.}:
\begin{itemize}
\item \textbf{Normalized AVT:} The Average Visualization Time per epoch, our core metric for real-time capability.
\item \textbf{Normalized Max/Average Delay:} The total delay in terms of training epochs.
\end{itemize}

\noindent\textbf{Results.}
The efficiency comparison, presented in Table~\ref{tb:efficiency}, demonstrates \tool's profound efficiency advantage over existing methods. 
The normalized results clearly show that both TimeVis and DVI are impractical for live monitoring, with delays often accumulating to hundreds or even thousands of equivalent training epochs. 
DVI's sequential design, in particular, leads to rapidly accumulating delays, while TimeVis is a purely post-hoc tool.
In stark contrast, \tool's efficiency enables true live-update visualization. 
For all ViT-based models and the computationally intensive CNN-based Food-101 task, \tool's AVT/ATT ratio is less than 1.0. 
It proves \tool{} can generate the visualization for a given checkpoint before the next checkpoint is even created, achieving a true real-time feedback loop with zero accumulating delay.
For the faster CIFAR-CNN tasks, \tool{} incurs a minimal average delay of only 3-6 epochs. 
Given that these models are trained for hundreds of epochs, this small, constant-time overhead is entirely acceptable and provides a near-live view of the training dynamics. This practical, live-update capability is essential for enabling timely interventions and saving computational resources—a feat not achievable by the baseline methods.

\subsection{RQ2: Visualization Quality}
We evaluate visualization quality quantitatively through three standard metrics and qualitatively by assessing the interpretability of the resulting visualizations.

\noindent\textbf{Quantitative Evaluation Metrics.}
Following established work~\cite{gigante2019visualizing,yang2022deepvisualinsight,yang2022temporality}, we assess performance using three key metrics covering spatial and temporal fidelity. Formal definitions are in Appendix~\ref{sec:visual-performace}.
\begin{itemize}
    % \item \textbf{Trustworthiness}: Measures the extent to which the local structure of the high-dimensional data is preserved in the low-dimensional visualization~\cite{pmlr-v5-maaten09a}.
    \item \textbf{Intraslice Neighbor Preservation}: Evaluates how well the $k$-nearest neighbors within a single time step are preserved after dimensionality reduction.
    \item \textbf{Reconstruction Accuracy}: Measures whether the reconstructed representations maintain the same predictions as the original high-dimensional data.
    \item \textbf{Interslice Neighbor Ranking Correlation}: Assesses whether the visualization accurately captures the movement of samples across different time steps by comparing the ranking of distances in high-dimensional and low-dimensional spaces.
\end{itemize}
The formal definitions of those metrics are in Appendix~\ref{sec:visual-performace}.

\noindent\textbf{Quantitative Results.}
As the post-hoc baselines represent a practical upper bound on visualization quality, our goal is to demonstrate that \tool{} achieves comparable fidelity while operating in a live, real-time setting. 
For spatial properties, \tool{} consistently matches or exceeds the performance of baseline methods in Intraslice Neighbor Preservation across all training stages, as shown in Figure \ref{fig:nn-resnet}.
Results for the ViT architecture, which lead to a similar conclusion, are in Appendix~\ref{sec:exp-results}.
Figure~\ref{fig:ppr-resnet} shows the Reconstruction Accuracy, where \tool's advantage becomes more pronounced.
It again delivers state-of-the-art performance while demonstrating superior robustness. 
Notably, TimeVis fails entirely on the ResNet-FOOD101 task, and DVI fails on the ViT-CIFAR100 task. 
In contrast, \tool{} performed reliably across all tested configurations, showing particular strength on more complex datasets like FOOD101.

For temporal properties, Figure~\ref{fig:tr-resnet} shows that \tool's Interslice Neighbor Ranking Correlation is highly competitive with TimeVis and significantly outperforms DVI. 
This is a strong result, as TimeVis processes all time steps at once, establishing a practical upper bound on performance. 
\tool{} achieves this comparable performance by efficiently sampling from variously spaced time intervals, striking a balance between historical context and computational feasibility.
The single exception is DVI's higher temporal score on ResNet50-Food101. However, this superior score is a misleading artifact from a pathological visualization where the data collapses into a line.
\begin{figure}[t!]
    \centering
    \begin{subfigure}[]{0.45\textwidth}
        \includegraphics[width=\textwidth]{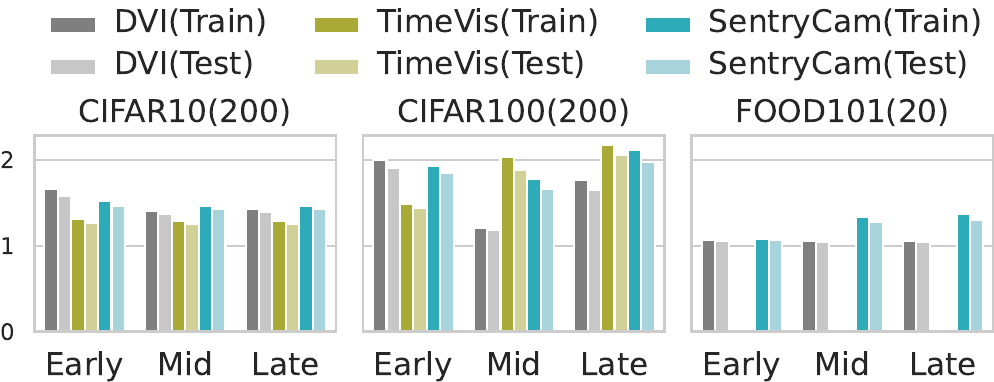}
        \caption{Intraslice Neighbor Preservation ($k=15$)}
        \label{fig:nn-resnet}
    \end{subfigure}
    \begin{subfigure}[]{0.45\textwidth}
        \includegraphics[width=\textwidth]{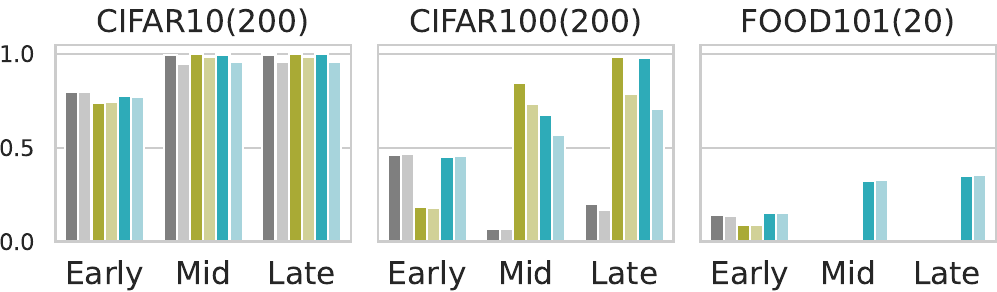}
        \caption{Reconstruction Accuracy}
        \label{fig:ppr-resnet}
    \end{subfigure}
    \begin{subfigure}[]{0.45\textwidth}
        \includegraphics[width=\textwidth]{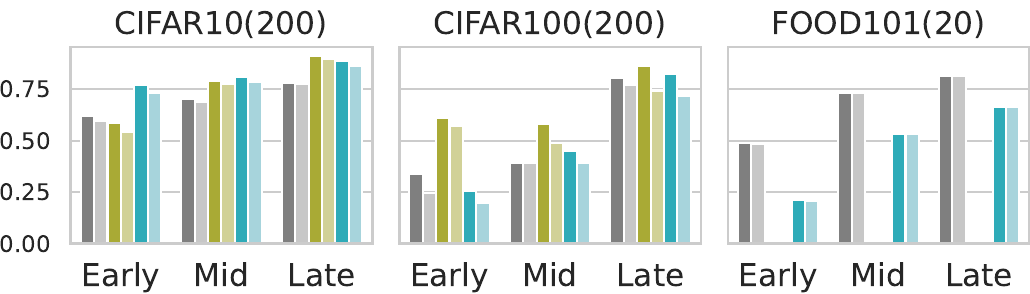}
        \caption{Interslice Neighbor Ranking Correlation}
        \label{fig:tr-resnet}
        \end{subfigure}
    \caption{\textbf{Visualization Performance of ResNet Architecture.}}
\end{figure}

\begin{figure}[t!]
     \centering
     \begin{subfigure}[b]{0.13\textwidth}
         \centering
         \includegraphics[width=\textwidth]{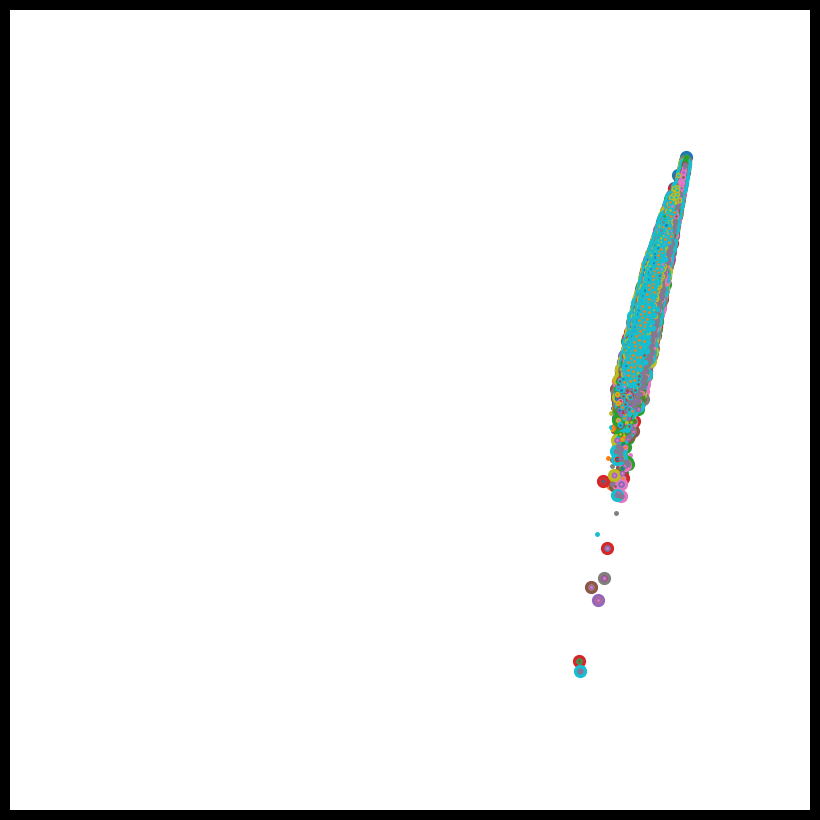}
         \caption{DVI}
         \label{fig:resnet_food101_dvi_20}
     \end{subfigure}
     \hspace{6mm}
     \begin{subfigure}[b]{0.13\textwidth}
         \centering
         \includegraphics[width=\textwidth]{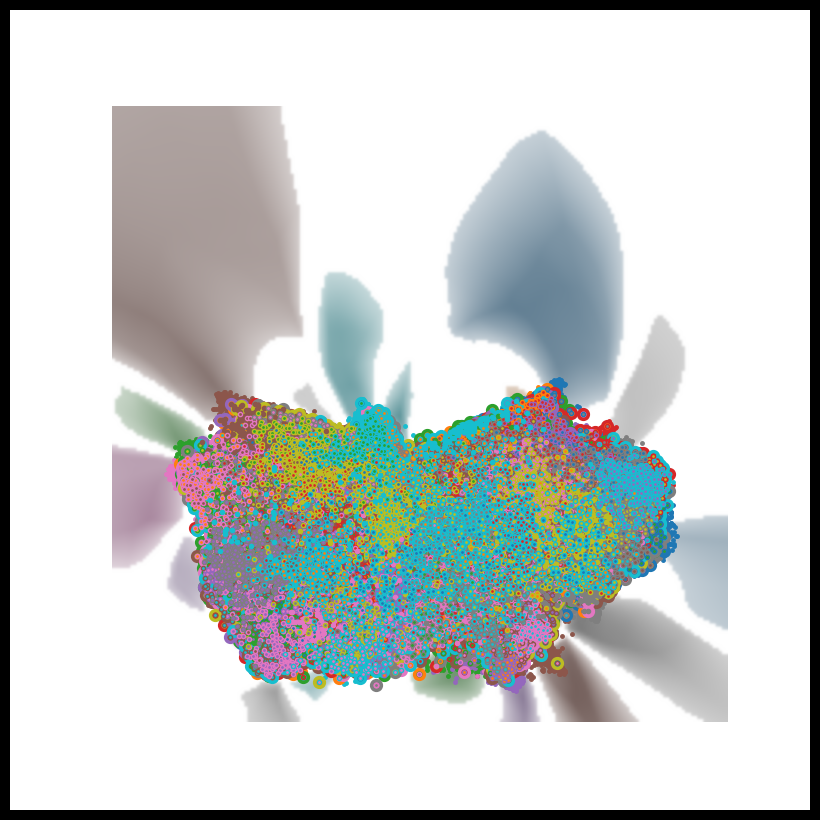}
         \caption{\tool}
         \label{fig:resnet_food101_tdvi_20}
     \end{subfigure}
     \caption{\textbf{Comparasion of visualization results on ResNet over FOOD101 dataset.} The representations are shown with dots with color being their labels. The color of the background represents prediction and the white part is the decision boundary. For example, a blue sample lies in the red region indicates that a sample belonging to the blue class is misclassified by the model as red class. Note how \tool{} (b) maintains distinct, interpretable clusters for different classes, while DVI (a) collapses the data into a single pathological structure, obscuring class relationships.}
    \label{fig:resnet-food101}
\end{figure}

\noindent\textbf{Qualitative Analysis.}\label{sec:exp-quality}
While metrics are important, a visualization's ultimate value lies in its interpretability. 
As an illustrative case study, Figure~\ref{fig:resnet-food101} compares visualizations on the challenging Food-101 task (further qualitative results are in Appendix~\ref{app:vf}). The figure highlights a scenario where a baseline's pathological output could mislead a practitioner. While TimeVis failed to generate a valid output, DVI collapses the data into a single, elongated structure. This artifact misleadingly inflates its temporal score but renders it useless for auditing class separation. In contrast, \tool{} produces a coherent and interpretable classification landscape, maintaining the integrity of the data clusters and providing a more balanced and faithful visualization.

\subsection{RQ3: Practical Utility for Real-Time Auditing}
While post-hoc analysis with the visualization trajectories is valuable for tasks like data anomaly detection and model debugging~\cite{yang2023deepdebugger}, a key challenge is whether a live-update visualization can proactively audit the training process. 
A successful auditing tool must provide alerts for impending failures ahead of conventional metrics, thereby saving significant computational resources.
To evaluate \tool's utility in this capacity, we design a scenario to diagnose training instability in real-time.

\noindent\textbf{Task Setup}
We train a ResNet-34 on the CIFAR-100 dataset using standard hyperparameters (SGD optimizer, MultiStepLR scheduler, batch size 128) but set an intentionally high learning rate of $0.7$.
This setup is engineered to induce training instability, where the model initially converges before catastrophically diverging. 
% While our analysis focuses on the geometric metrics of representation collapse and explosion, \tool{} is an extensible framework that allows developers to implement custom alerts tailored to other specific training failure scenarios.

Note that the geometry-based alert conditions for instability are just one example; developers can readily implement customized triggers based on different geometric properties or training scenarios, making \tool{} a robust and flexible framework for model auditing.

\noindent\textbf{Alert Conditions.}
To quantitatively compare monitoring methods, we define specific alert conditions for both conventional and representation-based metrics.
\begin{itemize}
    \item \textbf{Conventional Alert}: The baseline alert is triggered when the smoothed validation loss shows a sustained increase for at least k=2 consecutive epochs. This signals that the model's overall performance has begun to degrade.
    \item \textbf{\tool{} Alert}: The diagnostic alert is triggered by a sustained degradation in the model's geometric health state. We define this as the earliest epoch where either the smoothed Inter-Cluster Distance (separation) decreases or the smoothed Intra-Cluster Variance (cohesion) increases for k=2 consecutive epochs. This two-part condition is designed to detect either representation collapse or explosion. A detailed description of the alert's margin-based significance check is provided in Appendix~\ref{sec:alert-condition}.
\end{itemize}
We hypothesize that the \tool{} alert, by monitoring the internal representation geometry, will trigger earlier than the loss-based alert, therefore provide a timely warning of irreversible training failure.

\noindent\textbf{Results}
\begin{figure}
    \centering
    \includegraphics[width=\linewidth]{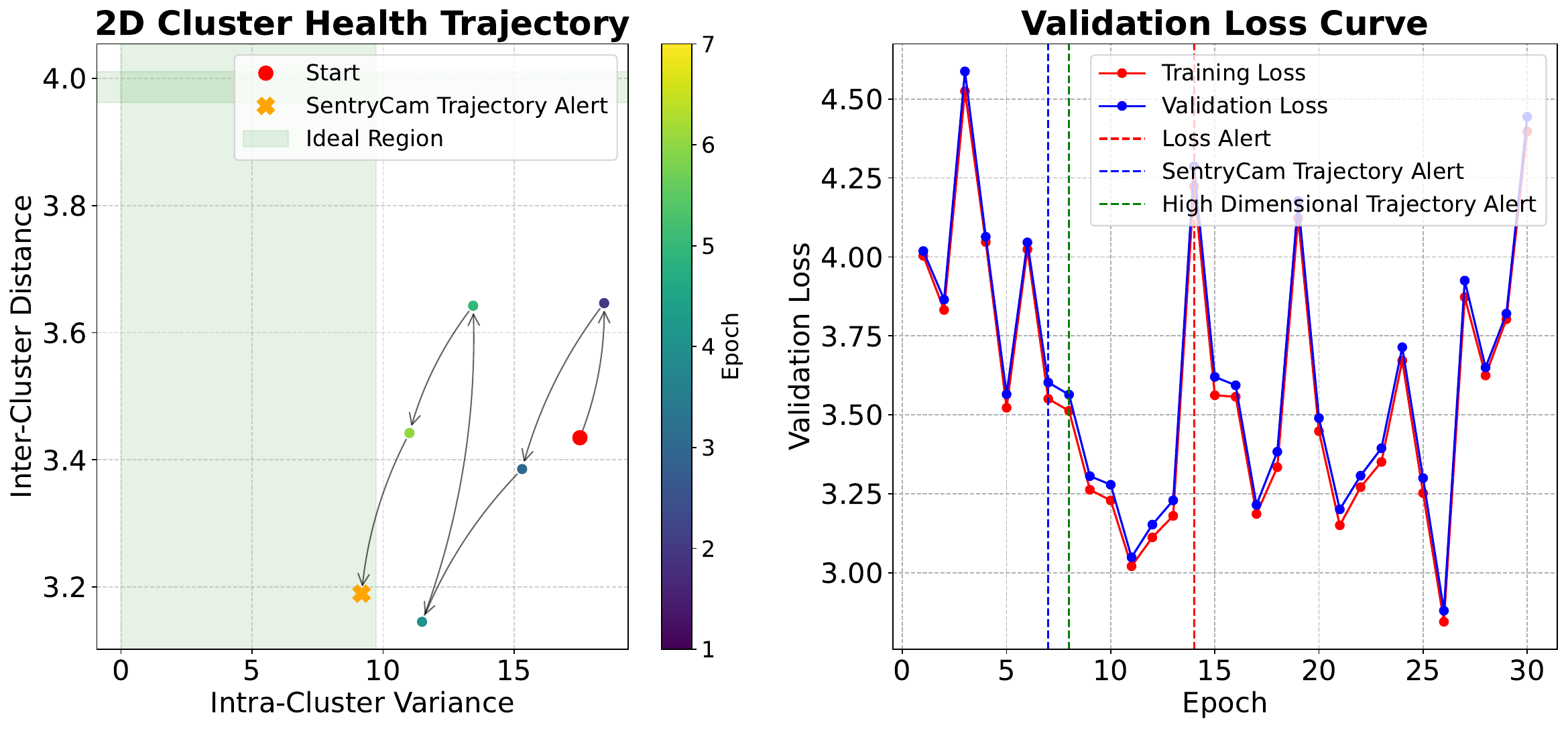}
    \caption{\textbf{\tool~Diagnostic for an Unstable Training Scenario.} 
(Left) A 2D Cluster Health Trajectory tracks the geometry of the latent space, plotting Inter-Cluster Distance (separation) against Intra-Cluster Variance (cohesion). The trajectory initially improves towards the ideal top-left corner before a sharp collapse. The green ``Ideal Region'' is defined by the top 20\% of metric values achieved during training. 
(Right) Corresponding loss curves show that the \tool~alert provides a significant early warning, triggering 7 epochs before the conventional validation loss alert.
}
    \label{fig:unstable-alert}
\end{figure}
Figure~\ref{fig:unstable-alert} confirm our hypothesis. 
The 2D Cluster Health Trajectory (left panel) provides a clear narrative of the model's failure. Initially, the trajectory moves towards the ``Ideal Region'' (top-left) as the model learns. 
However, at epoch 7, the trajectory takes a sharp downward turn, indicating a collapse in Inter-Cluster Distance. 
This triggers the \tool{} alert (orange X). 
As shown in the right panel, this \tool{} alert provides a significant early warning. 
It precedes the alert from the ground-truth high-dimensional trajectory (green, epoch 8) and occurs a full 7 epochs before the validation loss alert (red, epoch 14), which only fires after the model has already diverged and training has become irrecoverable.
Once the developers notice the alert, they can adjust the hyperparameters immediately to avoid further cost.

We further analyzed using the original high-dimensional activations directly as the trajectory for anomaly detection. 
We found this approach often provides a delayed alert, as seen in this experiment, or can be prone to false positives (see Appendix~\ref{sec:false-alarm}). 
We suspect this is because raw high-dimensional spaces are noisy and distances can be less meaningful due to the curse of dimensionality. This finding further strengthens the case for \tool, which not only provides visualization but also projects representations into a more stable and informative low-dimensional space for monitoring.

\section{Related Works}\label{sec:related-works}
\noindent\textbf{Visualizing Hidden Representations.} Analyzing the evolution of hidden activations is typically approached as a dimensionality reduction problem~\cite{telea2024seeing,gillmann2025extensible}. 
Non-parametric methods like t-SNE~\cite{van2008visualizing} and UMAP~\cite{mcinnes2018umap} excel at preserving local structure, while parametric methods like autoencoders~\cite{hinton2006reducing, moor2020topological} learn an explicit mapping function. 
Other techniques, such as Diffusion Maps~\cite{coifman2006diffusion} and PHATE~\cite{moon2017phate}, use spectral methods for relational preservation~\cite{jorgensen2025interpretablevisualizationsdataspaces}. 
Several methods have been specifically tailored to visualize the dynamics of DNN training, including M-PHATE~\cite{gigante2019visualizing}, DVI~\cite{yang2022deepvisualinsight}, and TimeVis~\cite{yang2022temporality}. 
While powerful for post-hoc analysis, these specialized tools lack the automation and real-time capabilities required for live monitoring, the primary focus of our work.

\noindent\textbf{Auditing with Training Dynamics.} A distinct line of research uses training dynamics to audit datasets and model behavior~\cite{badamasi2025exploring}.
These methods often track per-sample metrics over time to identify influential~\cite{kumar2024grokking}, ``forgettable''~\cite{toneva2018empirical}, or mislabeled~\cite{pleiss2020identifying} data points. Techniques like Data Maps~\cite{swayamdipta2020dataset} and Example Difficulty~\cite{baldock2021deep} characterize samples by their learning trajectory, while others monitor the evolution of neuron-level concepts~\cite{park2022conceptevo, khan2022demystifying}. 
These approaches are highly effective for targeted, data-centric auditing tasks. 
In contrast, \tool provides a holistic, global view of the entire representation space, enabling open-ended exploration and the detection of systemic geometric failures (like representation collapse) that are complementary to sample-specific analyses.

\section{Conclusion}\label{sec:conclusion}
In this work, we introduced \tool, a live-update visualization framework for the real-time auditing of deep neural network training. We established three core principles for practical monitoring—automation, live updates, and extensibility—and designed \tool{} to embody them through innovations in dynamic graph construction and stable projection learning.
Our experiments demonstrate that \tool{} achieves state-of-the-art visualization quality and efficiency, enabling true live monitoring where prior methods could not. Through case studies, we showcased its utility as a powerful auditing tool, providing early, actionable insights into training failures that are invisible to conventional metrics. By making the evolution of latent representations transparent and interpretable, \tool{} is a valuable tool for any deep learning practitioner seeking to build more robust and reliable models.

% We have presented \tool, a novel technique for visualizing the internal dynamics of deep neural network training. Our primary contribution is a framework built on three core principles for practical monitoring: automation, live updates, and extensibility. Through technical innovations in temporal memory, density-guided sampling, and hybrid normalization, \tool{} successfully overcomes the practical limitations of prior work.
% We demonstrated \tool's superiority in both visualization quality and computational efficiency across diverse experimental settings. Case studies in standard and continual learning further showcased its value, enabling the diagnosis of complex phenomena like training instability epochs before it was evident in performance metrics. In conclusion, this work establishes \tool{} as a valuable and practical tool for deep learning practitioners, making the crucial but often opaque process of model training more transparent and controllable.
{
    \small
    \bibliographystyle{ieeenat_fullname}
    \bibliography{main}
}

\clearpage
\setcounter{page}{1}
\maketitlesupplementary

\appendix

\renewcommand{\thesection}{\Alph{section}}
\renewcommand{\thesubsection}{\Alph{section}.\arabic{subsection}}
\onecolumn
\section{Background on Parametric UMAP}\label{app:background}

Parametric dimension reduction techniques employ parameterized networks (e.g., autoencoders) to project high-dimensional data into lower-dimensional spaces. 
% These methods are particularly advantageous compared to non-parametric approaches due to their efficiency in handling new, unseen data. 
In this work, we build upon parametric UMAP~\cite{sainburg2021parametric} for its proven ability to reduce dimensionality while faithfully preserving the topological structure of high-dimensional datasets.

The core of the visualization is a two-stage process: first, constructing a weighted graph representing the relationships in the high-dimensional space, and second, optimizing a low-dimensional embedding to match this structure.

\paragraph{1. High-Dimensional Graph Construction.}
Given two representations $\mathbf{x}_i, \mathbf{x}_j \in \mathbf{X}$ with an arbitrary distance metric $\mathrm{d}:\mathbb{R}^h \times \mathbb{R}^h\rightarrow \mathbb{R}_{\geq 0}$, we first computes a weighted edge $p_{ij}$ between them. 
This is done by calculating an asymmetric similarity $p_{j|i}$ based on the distance $\mathit{d}(\mathbf{x}_i, \mathbf{x}_j)$ and a local scaling factor $\sigma_i$ (Eq.~\ref{eq:pi2j}), and then symmetrizing the result using a probabilistic t-conorm (Eq.~\ref{eq:pij}):
\begin{equation}\label{eq:pi2j}
  p_{j|i} := \exp{\bigg({-\frac{d(\mathbf{x}_i, \mathbf{x}_j) - \rho_i}{\sigma_i}}\bigg)}
\end{equation}
where $\rho_i$ is the distance from $\mathbf{x}_i$ to its nearest neighbor.
\begin{equation}\label{eq:pij}
  p_{ij} = p_{i|j} + p_{j|i} - p_{i|j} \cdot p_{j|i}
\end{equation}
The resulting set of all $p_{ij}$ values forms a weighted graph, often called a fuzzy simplicial complex~\cite{mcinnes2018umap}, that captures the topological structure of the original data.

\paragraph{2. Low-Dimensional Structure Preservation.}
Next, for a set of corresponding low-dimensional points $\mathbf{Y} = \{\mathbf{y}_1, \dots, \mathbf{y}_N\}$ in the low-dimensional space $\mathbb{R}^l$, a similar set of weights $q_{ij}$ is calculated based on their distances (Eq.~\ref{eq:qij}). 
A parametric model (such as an autoencoder that maps from $\mathbf{X}$ to $\mathbf{Y}$) is then trained to minimize the cross-entropy between the high- and low-dimensional distributions, as formulated in the UMAP cost function (Eq.~\ref{eq:umap}):
\begin{equation}\label{eq:qij}
q_{ij} := \frac{1}{(1+a \Vert \mathbf{y}_i - \mathbf{y}_j\Vert^{2})^b}
\end{equation}
where $a$ and $b$ are learnable parameters of the low-dimensional embedding.
\begin{equation}\label{eq:umap}
\small
  \mathcal{C}_{umap} := \sum_{i} \sum_{j} \ \ \underbrace{p_{ij}\cdot \log\bigg(\frac{p_{ij}}{q_{ij}}\bigg)}_{\text{Attraction}} + \underbrace{(1-p_{ij})\cdot \log\bigg(\frac{1-p_{ij}}{1-q_{ij}}\bigg)}_{\text{Repulsion}}
\end{equation}
This cost function preserves the topological structure of the high-dimensional data by exerting an attractive force on similar point pairs (high $p_{ij}$) and a repulsive force on dissimilar ones (low $p_{ij}$), effectively arranging the low-dimensional embedding to reconstruct the original graph.

\section{Experiment Configuration Details}
\subsection{Datasets for Visualization}\label{sec:datasets}
The details of the image classification datasets are as follows:
\begin{center}
\begin{table}[h]
\centering

\label{tb:dataset}
\begin{tabular}{|c|c|c|c|c|c|}
\toprule
Datasets & Classes & Image Size & Train Size & Test Size & Num per Classes \\ \hline
CIFAR10  & 10      & 32         & 50000      & 10000     & 5000            \\ 
CIFAR100 & 100     & 32         & 50000      & 10000     & 500             \\ 
FOOD101  & 101     & 224        & 75750      & 25250     & 750             \\ 
\bottomrule
\end{tabular}
\caption{Datasets details for visualization}
\end{table}
\end{center}

\subsection{Subject Model Settings}\label{sec:subject-models}
The selection of training settings including model architecture, optimizer and batch size are shown in Table \ref{tb:subject-models}.
% \begin{sidewaystable}
\begin{table}[h!]
% \footnotesize	
% \scriptsize
\centering
\resizebox{0.95\columnwidth}{!}{

\begin{tabular}{|c|c|c|c|c|c|c|c|c|}
\toprule
Training Configuration                    & Dataset  & Model Arch                                                        & lr   & optimizer & batch size & scheduler         & epochs & Final Accu \\ \hline
\multirow{4}{*}{Train from Scratch} & CIFAR10  & ResNet18                                                          & 1e-2 & SGD       & 128        & MultiStepLR       & 200    & 0.9393     \\ \cline{2-9} 
                                    & CIFAR100 & ResNet34                                                          & 1e-2 & SGD       & 128        & MultiStepLR       & 200    & 0.7721     \\ \cline{2-9} 
                                    & CIFAR10  & \begin{tabular}[c]{@{}c@{}}ViT\\ (8 heads, 6 layers)\end{tabular} & 1e-4 & Adam      & 128        & CosineAnnealingLR & 200    & 0.8035     \\ \cline{2-9} 
                                    & CIFAR100 & \begin{tabular}[c]{@{}c@{}}ViT\\ (8 heads, 6 layers)\end{tabular} & 1e-4 & Adam      & 128        & CosineAnnealingLR & 200    & 0.5581     \\ \hline
\multirow{2}{*}{Fine-Tune}          & FOOD101  & ResNet50                                                          & 1e-4 & Adam      & 256        & MultiStepLR       & 20     & 0.6723     \\ \cline{2-9} 
                                    & FOOD101  & ViT/B-16                                                          & 3e-4 & Adam      & 256        & CosineAnnealingLR & 20     & 0.8328     \\
\bottomrule
\end{tabular}
}
\caption{Training Configurations for subject models}
\label{tb:subject-models}
\end{table}
% \end{sidewaystable}

\subsection{Visualization Performance Metrics}\label{sec:visual-performace}
We evaluate the visualization performance quantitatively with the following metrics:
\begin{itemize}
    % \item \textbf{Trustworthiness}: evaluate to what extent the local structure is retained. Formally, 
    % \begin{equation}
    %     T(k) = 1-\frac{2}{Nk(2N-3k-1)\sum_{i=1}^{N}}\sum_{j \in \mathcal{N}_i^k} \max(0, (r(I,j)-k))
    % \end{equation}
    % where $N$ and $k$ represent the number of samples and the neighborhood strength respectively~\cite{pmlr-v5-maaten09a}.
    \item \textbf{Intraslice Neighbor Preservation}: evaluate how many k nearest neighbors are preserved after dimension reduction at any time step $t$.
    \begin{equation}
        \frac{1}{N}\sum^{i=1}_{N}\bigg|\mathcal{N}^k_{High} (t, i)\cap \mathcal{N}^k_{Low}(t, i)\bigg|
    \end{equation}
    where $N$ is the total number of samples, and $\mathcal{N}^k_{High} (t, i)$ is the k nearest neighbors of sample $i$ at time step $t$ from High dimensional space.
    % \item \textbf{Interslice Neighbor Preservation}: evaluation of k nearest neighbor preserving rate of a sample across all checkpoints.
    % \begin{equation}
    %     \frac{1}{k}\bigg|\mathcal{N}^k_{High} (:, i)\cap \mathcal{N}^k_{Low}(:, i)\bigg|
    % \end{equation}
    % where $N$ is the total number of samples, and $\mathcal{N}^k_{High} (t, i)$ is the k nearest neighbors of sample $i$ at time step $t$ from High dimensional space.
    % \item  \textbf{Interslice Loss Correlation}: evaluation of the Spearman correlation of the rate of change of each low dimensional embedding with the rate of change of the loss.
    % \begin{equation}
    %     Corr_i = Spearman(\{ \Delta L_i \} , \{ \Delta d_i \} )
    % \end{equation}
    % where $\{ \Delta L_i \} = \{ L^{t+1}_i - L^t_i \}_{t=1}^{T-1}$ is the rate of change in loss within timesteps $T$ for sample $i$, and $\{ \Delta d_i \}=\biggl\{ \sqrt{(y^{t+1}_i-y^t_i)^2}\biggr\}_{t=1}^{T-1}$ is the rate of change of position of sample $i$ in low dimensional space.    
    \item \textbf{Reconstruction Accuracy}: evaluate whether the reconstructed representations have the same prediction as the original ones.

    \item \textbf{Interslice Neighbor Ranking Correlation:} evaluate whether the visualization method could faithfully show the movement of the samples across different time steps.
    Formally, for a sample $x^t_i$ at time step $t$, let $r^t_i$ and $\tilde{r}^t_i$ be the ranking of itself in other time steps $\{x^t_i\}_{t\in[T]}$ ordered by the distance in high-dimensional space and low-dimensional space respectively, we define the correlation between the two rankings as~\cite{yang2022temporality}.:
    \begin{equation}
        Corr_i = spearman(r^t_i, \tilde{r}^t_i)
    \end{equation}
\end{itemize}

\section{Additional Experiment Results}\label{sec:exp-results}
In this section, we show the full results of visualization performance in section~\ref{ap:vp} and visualization figures in section~\ref{app:vf}.
\subsection{Visualization Performance}\label{ap:vp}
\begin{figure}[h!]
     \centering
     \begin{subfigure}[b]{0.45\textwidth}
         \centering
         \includegraphics[width=\textwidth]{exp-results/nn_ResNet_15.pdf}
         \caption{ResNet}
         % \label{fig:nn-resnet}
     \end{subfigure}
     % \hfill
     \begin{subfigure}[b]{0.45\textwidth}
         \centering
         \includegraphics[width=\textwidth]{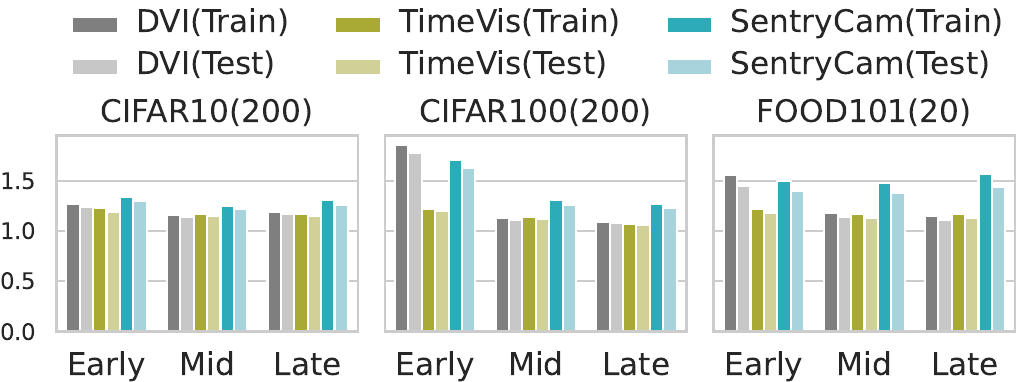}
         \caption{ViT}
         % \label{fig:nn-vit}
     \end{subfigure}
     \caption{Intraslice Neighbor Preservation (k=15)}
    \label{fig:nn}
\end{figure}

\begin{figure}[h!]
     \centering
     \begin{subfigure}[]{0.45\textwidth}
         \centering
         \includegraphics[width=\textwidth]{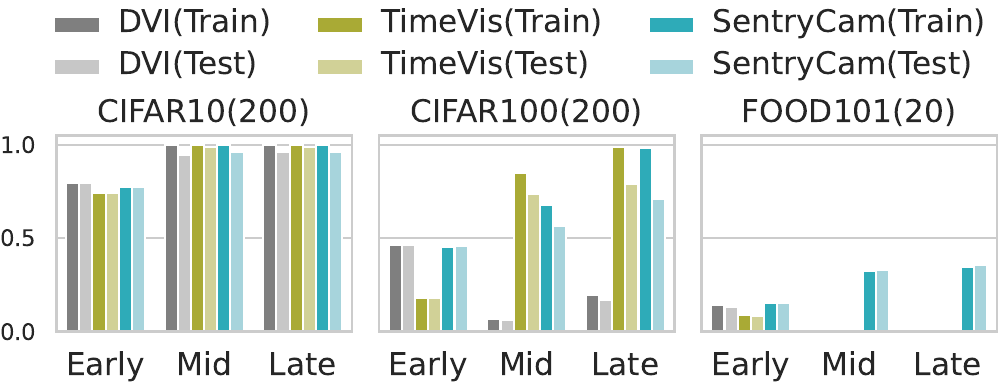}
         \caption{ResNet}
         % \label{fig:nn-resnet}
     \end{subfigure}
     % \hfill
     \begin{subfigure}[]{0.45\textwidth}
         \centering
         \includegraphics[width=\textwidth]{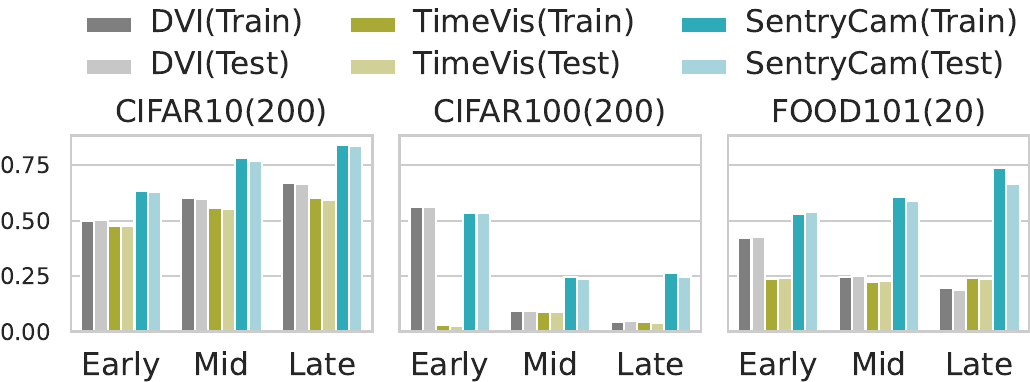}
         \caption{ViT}
         % \label{fig:nn-vit}
     \end{subfigure}
    \caption{Reconstruction Accuracy}
    \label{fig:recon}
\end{figure}

\begin{figure}[h!]
     \centering
     \begin{subfigure}[b]{0.45\textwidth}
         \centering
         \includegraphics[width=\textwidth]{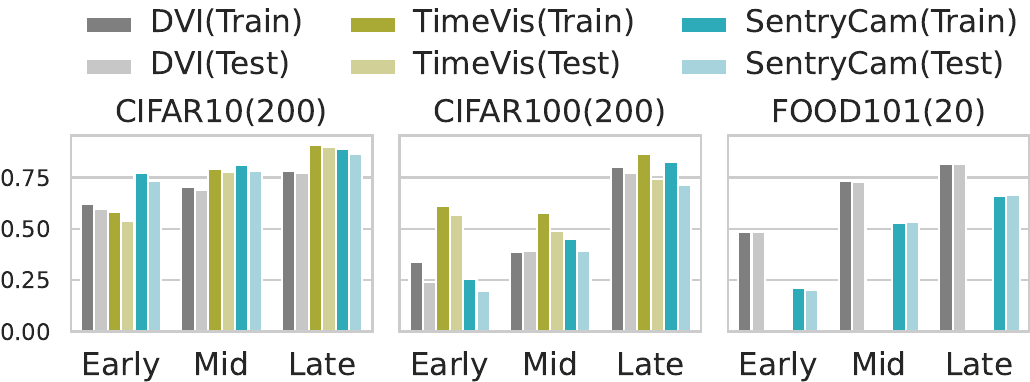}
         \caption{ResNet}
         % \label{fig:tr-resnet}
     \end{subfigure}
     % \hfill
     \begin{subfigure}[b]{0.45\textwidth}
         \centering
         \includegraphics[width=\textwidth]{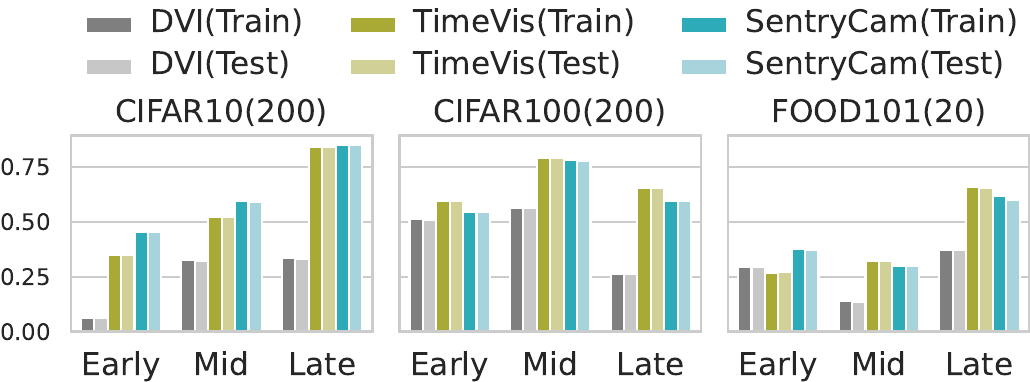}
         \caption{ViT}
         % \label{fig:tr-vit}
     \end{subfigure}
     \caption{Interslice Neighbor Ranking Correlation}
    \label{fig:tr}
\end{figure}

The visualization performance of different architectures is shown here.
The intraslice neighbor preservation is shown in Figure~\ref{fig:nn}; the reconstruction accuracy is shown in Figure~\ref{fig:recon}; and finally the interslice neighbor ranking correlation is shown in Figure~\ref{fig:tr}.

% \begin{figure}[h!]
%      \centering
%      \begin{subfigure}[]{0.45\textwidth}
%          \centering
%          \includegraphics[width=\textwidth]{exp-results/trustworthiness_ResNet_15.pdf}
%          \caption{ResNet}
%          % \label{fig:trustworthiness-resnet}
%      \end{subfigure}
%      % \hfill
%      \begin{subfigure}[]{0.45\textwidth}
%          \centering
%          \includegraphics[width=\textwidth]{exp-results/trustworthiness_ViT_15.pdf}
%          \caption{ViT}
%          % \label{fig:trustworthiness-vit}
%      \end{subfigure}
%      \caption{Intraslice Trustworthiness}
%     \label{fig:trustworthiness}
% \end{figure}

\subsection{Visualization Figure of FOOD101 with ViT Architecture}\label{app:vf}
\begin{figure*}[h!]
     \centering
     \begin{subfigure}[b]{0.3\textwidth}
         \centering
         \includegraphics[width=\textwidth]{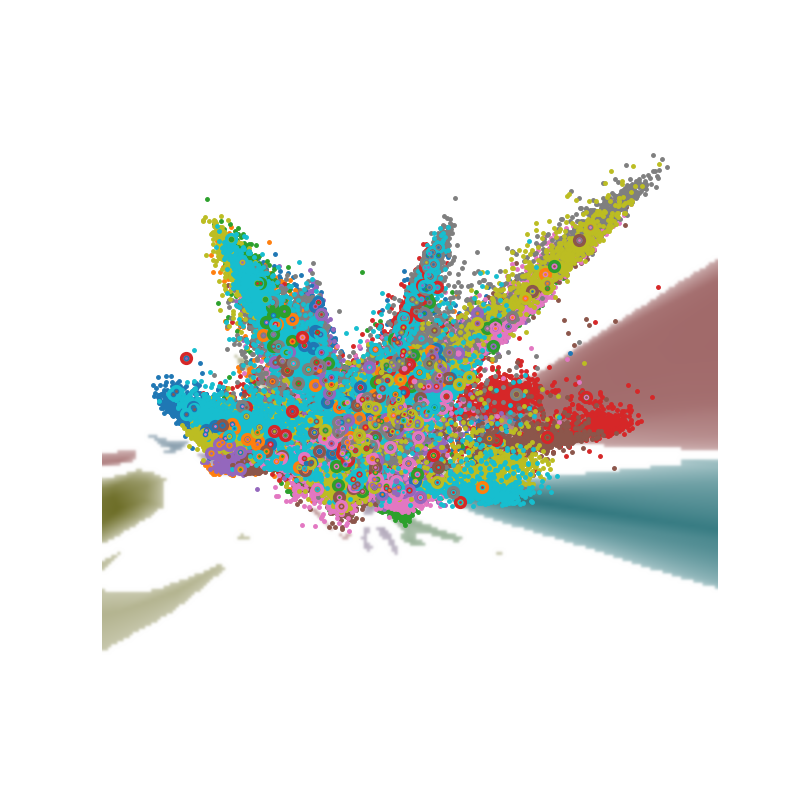}
         \caption{DVI}
         \label{fig:vit_food101_dvi_20}
     \end{subfigure}
     \hfill
     \begin{subfigure}[b]{0.3\textwidth}
         \centering
         \includegraphics[width=\textwidth]{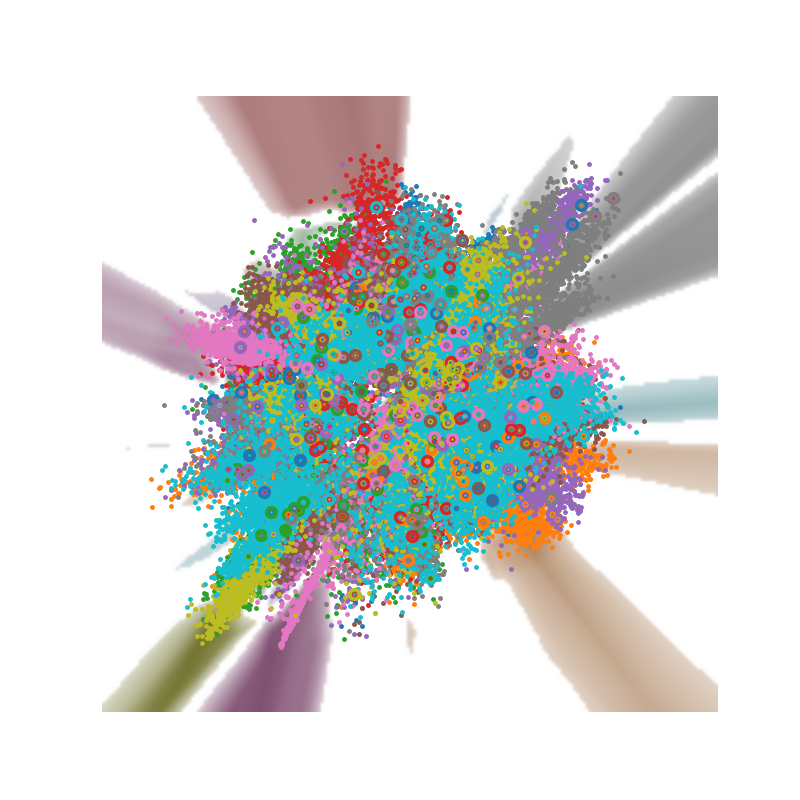}
         \caption{TimeVis}
         \label{fig:vit_food101_timevis_20}
     \end{subfigure}
     \hfill
     \begin{subfigure}[b]{0.3\textwidth}
         \centering
         \includegraphics[width=\textwidth]{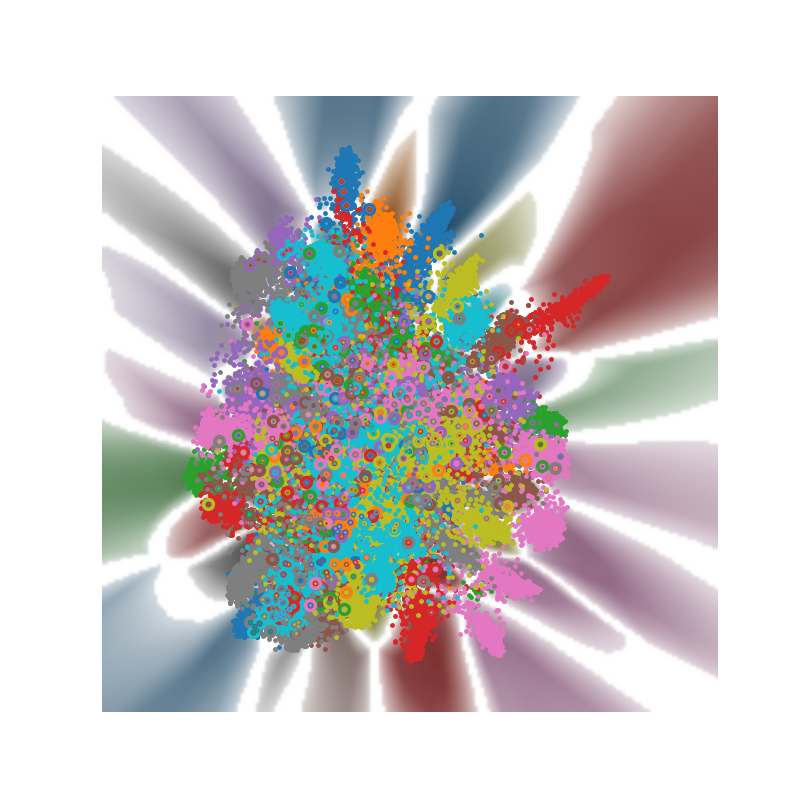}
         \caption{\tool}
         \label{fig:vit_food101_tdvi_20}
     \end{subfigure}
     \caption{Comparasion of visualization results on ViT over FOOD101 dataset. The white part represent decision boundary and the color of background represent classification prediction. Each sample is represented as a dot and the color of the dot represent its class label.}
    \label{fig:vit-food101}
\end{figure*}

We qualitatively evaluate the visualization results of ViT architecture.
In Figure~\ref{fig:vit-food101}, the visualization given by \tool can generate more meaningful decision boundary and more well-defined clusters, indicating the superior performance of \tool for hard dataset.

\section{RQ1: Raw wall-clock Time}\label{app:wall-clock}
% Normalized AVT/ATT is good, but add wall-clock AVT and ATT, plus GPU model, batch size, and memory footprint at runtime.
While the normalized AVT/ATT ratio presented in the main paper provides a hardware-agnostic measure of real-time feasibility, this section provides the absolute raw wall-clock times to give a complete picture of the computational costs.
\begin{table*}[ht]
\centering
\caption{Efficiency Comparison of Visualization Methods. AVT (Average Visualization Time), Max/Avg Delay, and ATT (Average Training Time per Epoch) are all reported in raw seconds.}
\label{tab:efficiency_raw_full}
\resizebox{\textwidth}{!}{
\begin{tabular}{@{}ll rrrr rrrr rrrr@{}}
\toprule
\multirow{2}{*}{\textbf{Model}} & \multirow{2}{*}{\textbf{Method}} & \multicolumn{4}{c}{\textbf{CIFAR-10}} & \multicolumn{4}{c}{\textbf{CIFAR-100}} & \multicolumn{4}{c}{\textbf{FOOD-101}} \\
\cmidrule(l){3-6} \cmidrule(l){7-10} \cmidrule(l){11-14}
& & ATT & AVT & Max Delay & Avg Delay & ATT & AVT & Max Delay & Avg Delay & ATT & AVT & Max Delay & Avg Delay \\
\midrule
\multirow{3}{*}{CNN based} & DVI       & 10.6 & 112.6 & 22521.6 & 12374.4 & 16.0 & 113.8 & 25974.6 & 14647.7 & 237.6 & 419.0 & 5247.3 & 5171.4 \\
                           & TimeVis   & 10.6 & -     & 5083.0  & 5083.3  & 16.0 & -     & 7179.5  & 7179.5  & 237.6 & -     & 8889.0 & 8889.0 \\
                           & SentryCam & 10.6 & 65.7  & 87.4    & 65.7    & 16.0 & 61.3  & 84.4    & 61.3    & 237.6 & 234.6 & 245.3  & 234.6  \\
\midrule
\multirow{3}{*}{ViT based} & DVI       & 27.8 & 120.3 & 24066.2 & 14863.4 & 50.6 & 119.4 & 23877.4 & 17058.3 & 683.2 & 677.5 & 14441.6& 14340.5 \\
                           & TimeVis   & 27.8 & -     & 8269.3  & 8269.3  & 50.6 & -     & 7254.1  & 7254.1  & 683.2 & -     & 8854.7 & 8854.7 \\
                           & SentryCam & 27.8 & 67.2  & 75.9    & 67.2    & 50.6 & 66.1  & 70.3    & 66.1    & 683.2 & 484.1 & 510.3  & 484.1  \\
\bottomrule
\end{tabular}
}
\end{table*}

\section{Optimal Sampling Ratio Search Algorithm}\label{sec:detailed-density-guided-sampling}
Algorithm~\ref{alg:density-guided-sampling} details our method for automatically determining the optimal data sampling ratio. The core of the algorithm is a \textbf{binary search} designed to efficiently find the highest possible sampling rate that maintains a data density above a specified threshold, $d_{th}$.

The process begins by defining a search space for the sampling ratio between 0 and 1. In each iteration, the algorithm samples a subset of the data, $D'$, at the midpoint \texttt{SampleRate} and calculates its density, $\delta'$. We define density as the inverse of the average distance to the k-nearest neighbors; thus, a higher density value is better. If the calculated density $\delta'$ is above the required threshold $d_{th}$, it indicates that the sample is sufficiently dense, and we can safely attempt a higher sampling rate in the next iteration by updating the \texttt{lower} bound. Conversely, if the density is below the threshold, the sample is too sparse, and we must reduce the sampling rate by updating the \texttt{upper} bound.

To ensure stability against random sampling variance, the density calculation at line 6 can be repeated multiple times (e.g., 3 times) with different random seeds, and the average value can be used. The search terminates when the interval between the upper and lower bounds is smaller than a predefined precision, $p$. The final \texttt{OptimalSampleRatio} represents the most aggressive sampling that can be applied without risking the loss of essential topological structures, thereby balancing computational efficiency and visualization fidelity.

As for the hyperparameter density threshold $d_{th}$, it can be set automatically to avoid manual tuning. A principled approach is to anchor it to the data's intrinsic density, for example, by setting $d_{th}$ to a fraction (e.g., 80\%) of the density calculated on the full, unsampled data at an early epoch. Empirically, we find the system is not highly sensitive to this value; due to the sharp phase transition shown in Figure~\ref{fig:sentrycam-empirical-observation}, any conservative threshold that keeps the sample on the stable plateau yields high-quality results.

\begin{algorithm}[h]
\caption{Optimal Sampling Ratio Search}
\label{alg:density-guided-sampling}
\begin{algorithmic}[1]
\REQUIRE $D$: Original dataset, $d_{th}$: Density threshold, $p$: Precision
\ENSURE $opt_r$: Optimal sampling ratio

\STATE $lower \leftarrow 0$, $upper \leftarrow 1$
\WHILE{$upper - lower > p$}
    \STATE $SampleRate \leftarrow (upper + lower) / 2$
    \STATE $D'$ = RandomSampling($D$, $SampleRate$)
    \STATE $\delta' \leftarrow$ CalculateDensity($D'$)
    \IF{$\delta' > d_{th}$}
        \STATE $lower \leftarrow SampleRate + p$
    \ELSE
        \STATE $upper \leftarrow SampleRate$
    \ENDIF
\ENDWHILE
\STATE $OptimalSampleRatio \leftarrow lower$
\RETURN $OptimalSampleRatio$
\end{algorithmic}
\end{algorithm}

\section{Empirical Observation of Pruning to Visualization Quality}\label{sec:empirical-ob}
\begin{figure}[h!]
     \centering
     \begin{subfigure}[b]{0.24\textwidth}
         \centering
         \includegraphics[width=\textwidth]{empirical-observation/relative_data_density.pdf}
         % \caption{Data density vs Sampling Ratio}
         \caption{}
         % \label{fig:sentrycam-pruning-density}
     \end{subfigure}
     % \hfill
     \begin{subfigure}[b]{0.3\textwidth}
         \centering
         \includegraphics[width=\textwidth]{empirical-observation/k_nearest_neighbor_preservation.pdf}
         % \caption{Visualization Performance}
         \caption{}
         % \label{fig:sentrycam-pruning-nn}
     \end{subfigure}
     % \hfill
     \begin{subfigure}[b]{0.3\textwidth}
         \centering
         \includegraphics[width=\textwidth]{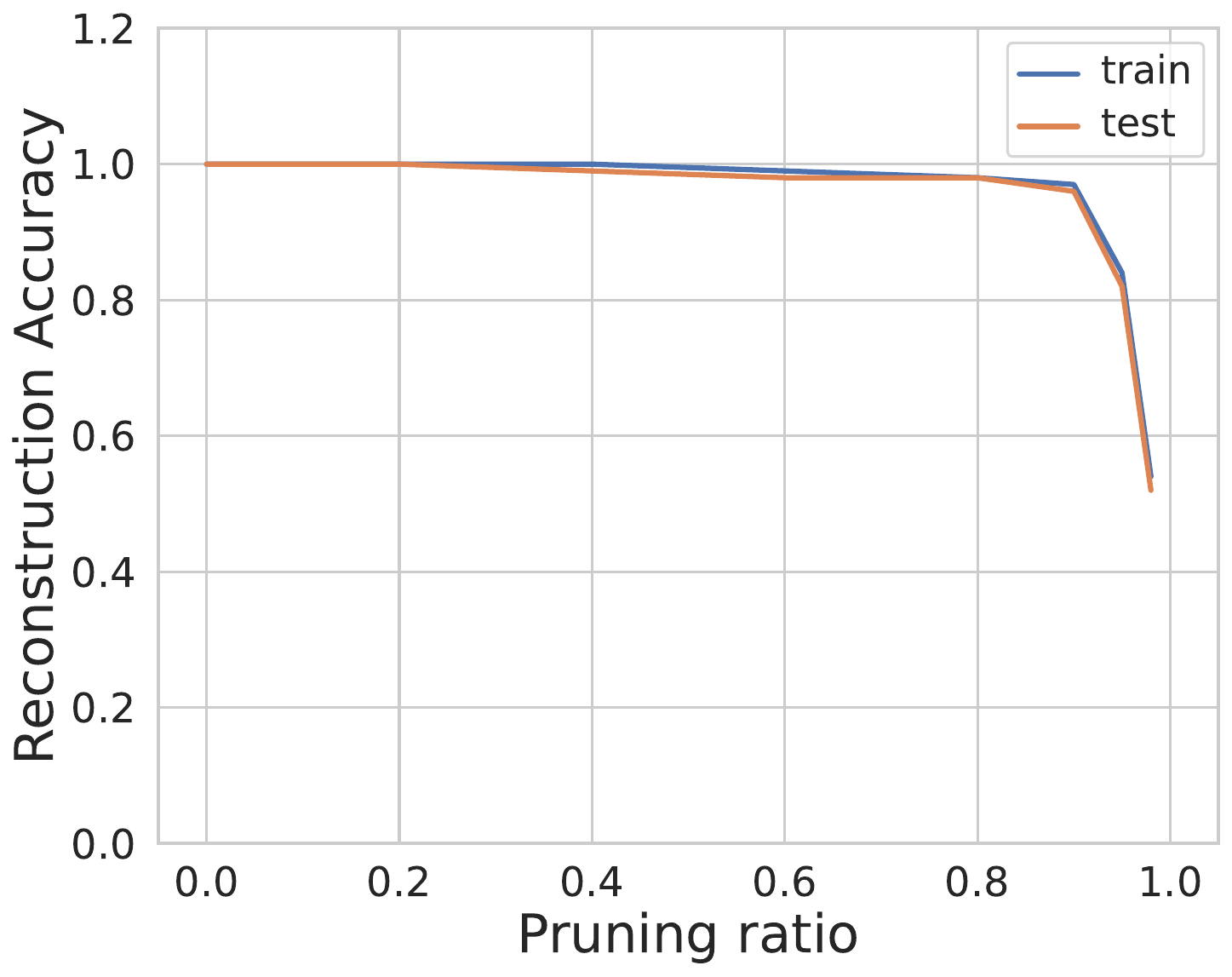}
         \caption{}
         % \label{fig:sentrycam-pruning-ppr}
     \end{subfigure}
     \caption{(a) The relationship between data pruning ratio and data relative density.
     (b) The relationship between the data pruning ratio and the nearest neighbor accuracy of dimension reduction using the pruned dataset.
     (c) The relationship between the data pruning ratio and the reconstruction accuracy of dimension reduction using the pruned dataset.
     }
    \label{fig:sentrycam-empirical-observation-full}
\end{figure}

To investigate the effect of pruning dataset to the visualization quality, following \ref{sec:visual-performace}, we evaluate the ntraslice Neighbor Preservation (k=15) and Reconstruction Accuracy of the visualization.
The relationship of pruning ratio and visualization performance is similar to the relationship of pruning ratio and relative data density between pruned dataset and original dataset.
The results are show in Figure~\ref{fig:sentrycam-empirical-observation-full}.

\section{Additional Results in RQ3}\label{sec:alert-condition}
\subsection{Alert Condition for Unstable Training}
In our case study of auditing training, beyond subjective visual inspection, we define a robust, quantitative trigger condition to automatically detect the onset of training anomalies. 
Our alerting mechanism is designed to be sensitive to meaningful changes while remaining resilient to stochastic noise inherent in the training process.
An alert is triggered for a given metric (e.g., Inter-Cluster Distance or Validation Loss) only when either two specific conditions are met:
\begin{itemize}
    \item \textbf{Sustained Trend Condition}: The metric must exhibit a degradation from its ideal state for a persistent period. We define this as a change in the "unhealthy" direction for at least $k$ consecutive epochs. For Inter-Cluster Distance, this is a decrease; for Intra-Cluster Variance and Validation Loss, this is an increase. This requirement ensures that the alert is not triggered by single-epoch fluctuations.
    \item \textbf{Significance Margin Condition}: The magnitude of the change must be statistically significant relative to the metric's recent behavior. We define a dynamic margin based on the metric's volatility. Specifically, the change in the most recent epoch, $|M(t) - M(t-1)|$, must exceed a fraction ($\alpha=0.25$) of the metric's standard deviation ($\sigma$) calculated over a sliding window of the last 10 epochs.
\end{itemize}

Therefore, the final trigger condition is: either a sustained negative trend for k epochs where the most recent change is greater than the dynamic margin $\alpha * \sigma$.
This two-part condition creates a highly reliable alert. The \tool alert is defined as the earliest epoch where either the Inter-Cluster Distance or the Intra-Cluster Variance metric meets this trigger condition. This is then compared against the trigger epoch derived from the validation loss to quantify the early-warning lead time provided by our representation-based analysis.

\subsection{Discussion on Auditing with High-Dimensional Representations}\label{sec:false-alarm}
A natural question arises: why not perform the geometric audit directly on the original, high-dimensional representations instead of their low-dimensional projections? While seemingly more direct, our experiments revealed that this approach is often less reliable, suffering from both delayed alerts for genuine failures and a susceptibility to false alarms during healthy training.

\begin{figure}[h]
\centering
\includegraphics[width=\linewidth]{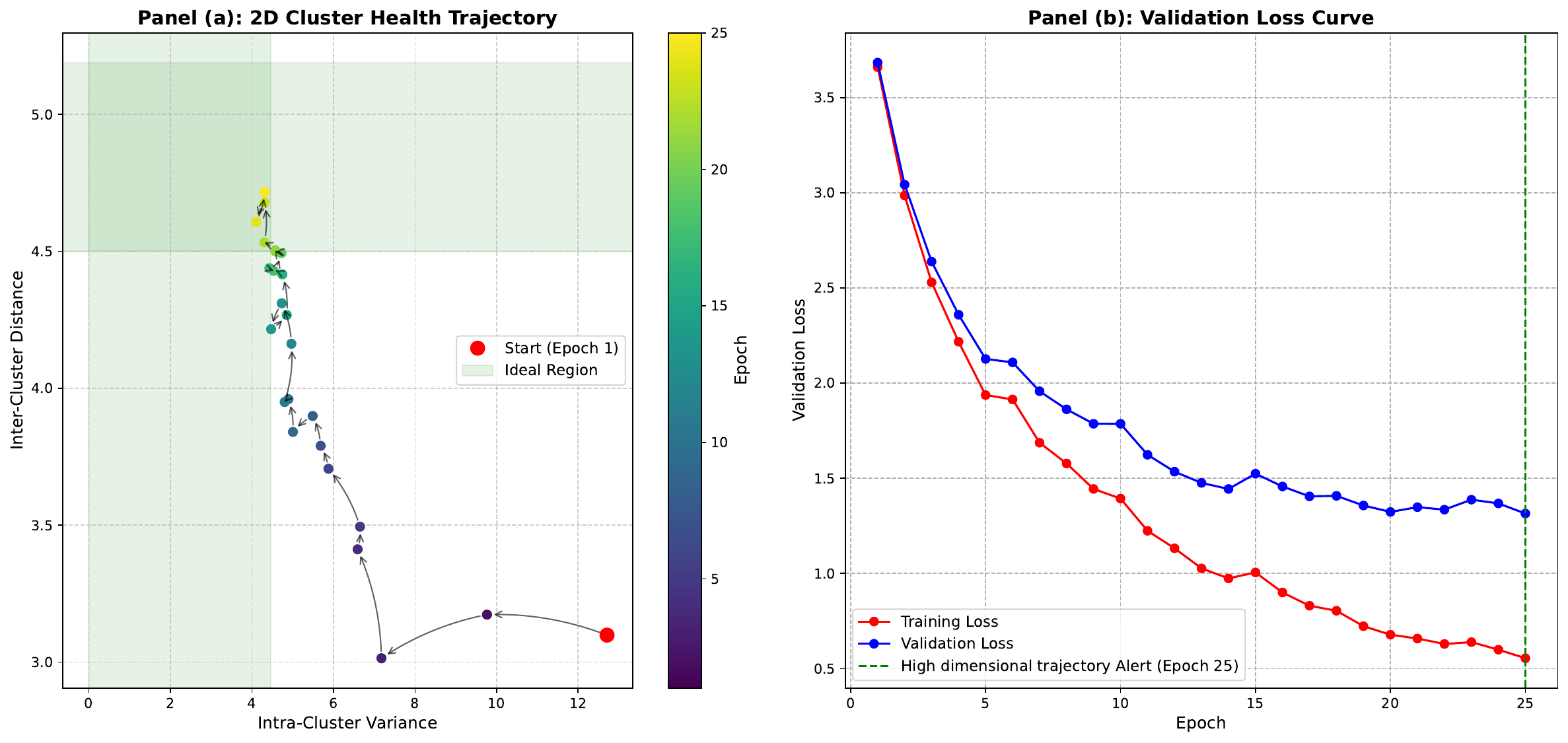}
\caption{\textbf{False Alarm from High-Dimensional Auditing during Healthy Training.} (a) \tool's 2D Cluster Health Trajectory shows a clear, monotonic improvement as the state moves towards the ideal top-left region. (b) The validation loss curve confirms this healthy convergence. Despite these positive indicators, the alert based on the raw high-dimensional trajectory incorrectly fires a false alarm at epoch 25.}
\label{fig:false-alarm}
\end{figure}

We observed a delayed alert in our training instability case study (Figure~\ref{fig:unstable-alert}), where the high-dimensional trajectory's alert (epoch 8) lagged behind \tool's (epoch 7). 
More critically, high-dimensional auditing can produce false alarms during stable, healthy training. 
Figure~\ref{fig:false-alarm} illustrates such a case during a standard ResNet-34 training on CIFAR-100. 
As shown in Panel (b), both the training and validation loss decrease steadily, indicating a healthy and convergent training process. 
\tool's 2D Cluster Health Trajectory in Panel (a) visually confirms this: the trajectory starts in a poor state (bottom-right) and moves consistently towards the ``Ideal Region'' (top-left), a clear sign of improving representation quality. However, the high-dimensional alert incorrectly triggers a false alarm at epoch 25, a point where the model is still clearly improving.

We hypothesize that these issues are rooted in the curse of dimensionality. In high-dimensional spaces, the concept of distance can become less meaningful; metrics based on Euclidean distance are often noisy and unstable as entire dimensions shift in scale and importance from one epoch to the next.
\tool's parametric projection model mitigates these issues by acting as a denoising and stabilizing filter. The projection learns to focus on the most salient geometric relationships while ignoring the high-frequency noise and irrelevant dimensional shifts that occur in the raw feature space. This results in cleaner, more reliable trajectories for the geometric audit metrics, leading to more timely and accurate alerts. Therefore, the visualization is not merely an interpretive aid but a crucial component for robust, automated auditing.

\section{Auditing Continual Learning with \tool}
To demonstrate \tool's utility in complex training scenarios, we apply it to Continual Learning (CL), a paradigm that addresses the challenge of adapting to a continuous influx of new data. 
We consider two common CL scenarios~\cite{vandeven2022three}: Domain-Incremental Learning (DIL), where the task remains the same but the data distribution shifts (e.g., classifying odd/even digits in different contexts of splitMNIST), and Class-Incremental Learning (CIL), where the model must learn new classes over time (e.g., classifying new digits as they are introduced).

\paragraph{Experimental Setup.}
We evaluate these scenarios using a CNN model on the splitMNIST dataset. We compare two canonical CL strategies: FROMP\cite{NEURIPS2020_2f3bbb97}, a weight regularization method, and Experience Replay (ER)\cite{NEURIPS2019_fa7cdfad}, a rehearsal-based method. 
The accuracy of each method across five sequential contexts is reported in Table~\ref{tb:cl-accu}. 
While these numerical results provide a high-level summary, \tool allows us to visualize the underlying geometric trade-offs that lead to these scores.
\begin{table}[h!]
\centering
\begin{tabular}{|c|c|c|c|c|c|c|c|}
\toprule
Scenario                & Baseline & Context 1 & Context 2 & Context 3 & Context 4 & Context 5 & Avg accu \\ \hline
\multirow{2}{*}{Domain} & ER       & 0.943     & 0.936     & 0.828     & 0.972     & 0.993     & 0.934    \\ \cline{2-8} 
                        & FROMP    & 0.608     & 0.931     & 0.607     & 0.977     & 0.993     & 0.823    \\ \hline
\multirow{2}{*}{Class}  & ER       & 0.931     & 0.823     & 0.764     & 0.895     & 0.988     & 0.880    \\ \cline{2-8} 
                        & FROMP    & 0.910     & 0.821     & 0.713     & 0.572     & 0.738     & 0.751    \\ 
\bottomrule
\end{tabular}
\caption{The accuracy for continual learning under different scenarios.}
\label{tb:cl-accu}
\end{table}

\begin{figure}[h!]
     \centering
     \begin{subfigure}[b]{0.15\textwidth}
         \centering
         \includegraphics[width=\textwidth]{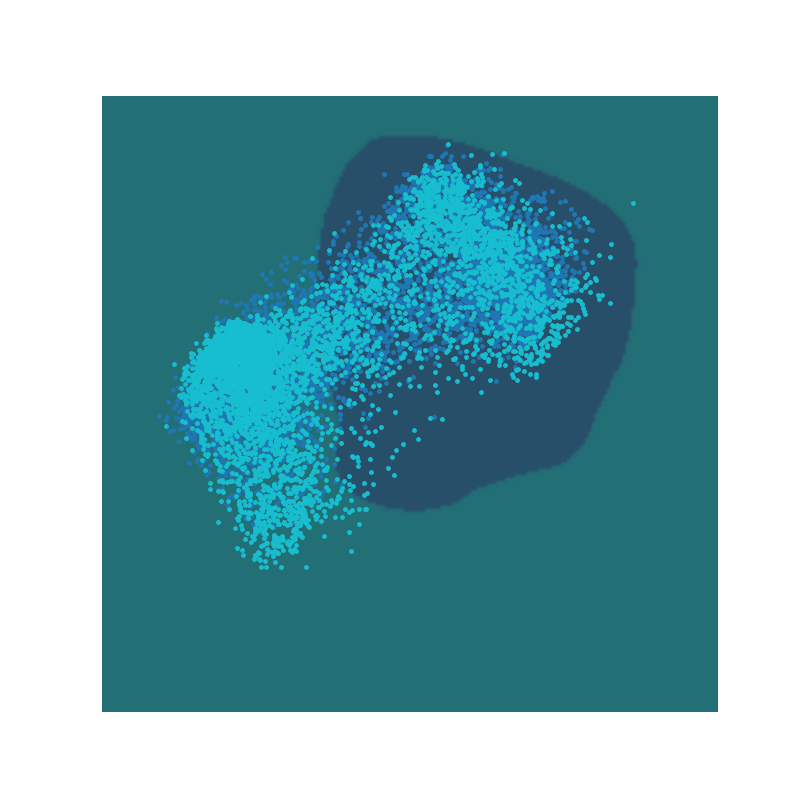}
         \caption{Context 1}
         \label{fig:cf_df_1}
     \end{subfigure}
     \begin{subfigure}[b]{0.15\textwidth}
         \centering
         \includegraphics[width=\textwidth]{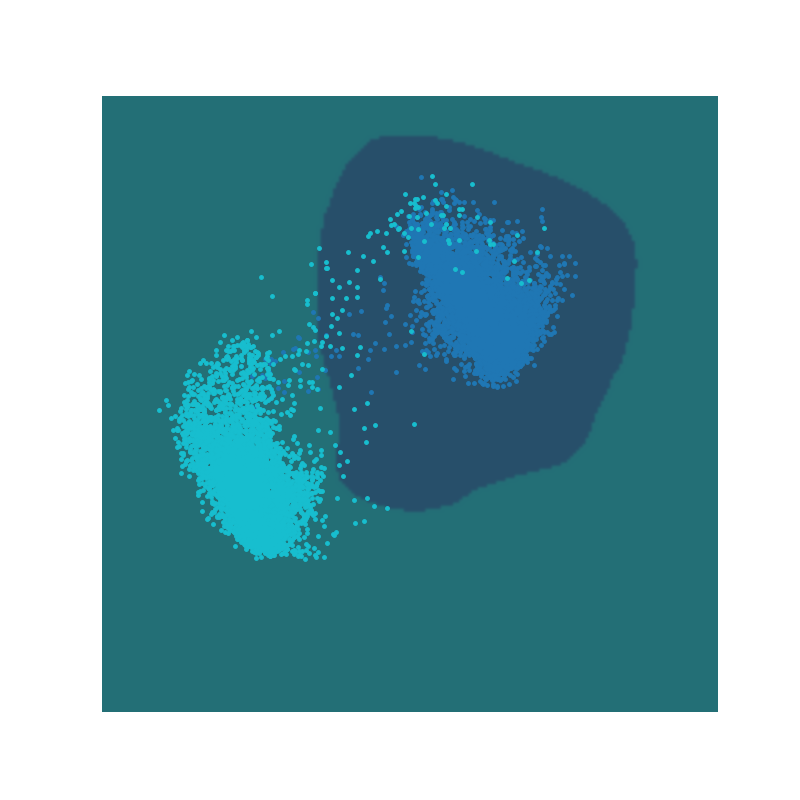}
         \caption{Context 2}
         \label{fig:cf_df_4}
     \end{subfigure}
     % \hfill
     \begin{subfigure}[b]{0.15\textwidth}
         \centering
         \includegraphics[width=\textwidth]{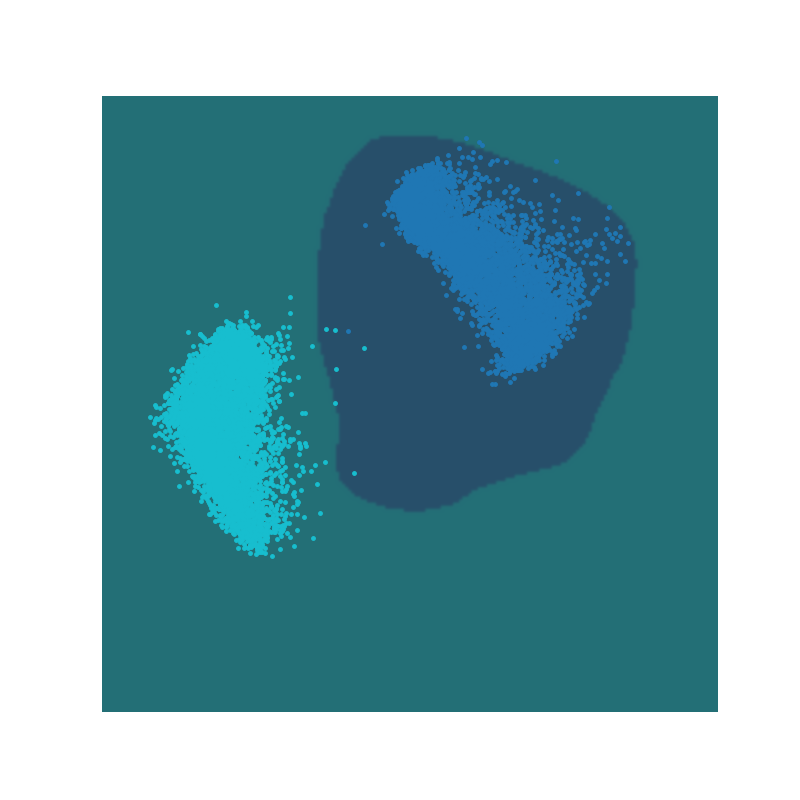}
         \caption{Context 3}
         \label{fig:cf_df_5}
     \end{subfigure}
     \caption{Visualization of data from previous contexts with the current decision boundary under the setting of Domain incremental learning with strategy FROMP. The lighter blue indicates odd class and darker blue indicates even class under different context.}
    \label{fig:cf-df}
\end{figure}

\paragraph{Visualizing Catastrophic Forgetting in DIL.}
Catastrophic forgetting occurs when a model discards prior knowledge while learning a new task. To visualize this, we use \tool to project data from earlier contexts onto the decision boundary of the final model trained with FROMP. As shown in Figure~\ref{fig:cf-df}, the representations for the earliest data (Context 1) have become highly disorganized and poorly separated. In contrast, data from more recent contexts are better-defined. This visualization makes the temporal nature of forgetting tangible: the older the knowledge, the more severely it is corrupted.

\paragraph{Visualizing Task Transfer and Plasticity in CIL.}
A key challenge in CIL is the plasticity-stability dilemma: the model must be plastic enough to learn new tasks (task transfer) but stable enough not to forget old ones. We use \tool to visualize the final feature space for both ER and FROMP.

\begin{figure}[h!]
     \centering
     \begin{subfigure}[b]{0.15\textwidth}
         \centering
         \includegraphics[width=\textwidth]{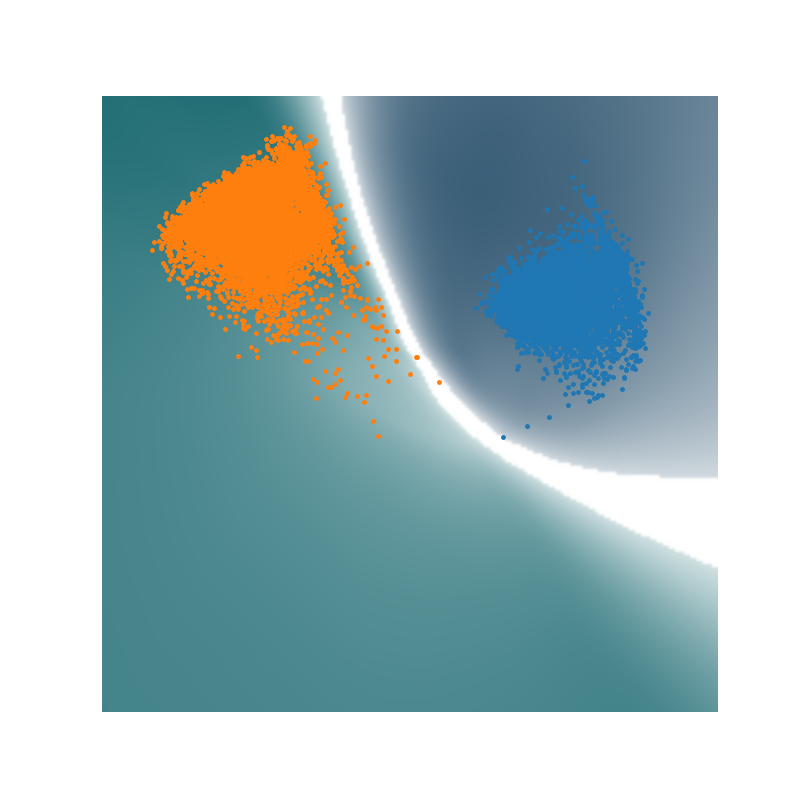}
         \caption{Context 1}
         \label{fig:tt_cr_1}
     \end{subfigure}
     \begin{subfigure}[b]{0.15\textwidth}
         \centering
         \includegraphics[width=\textwidth]{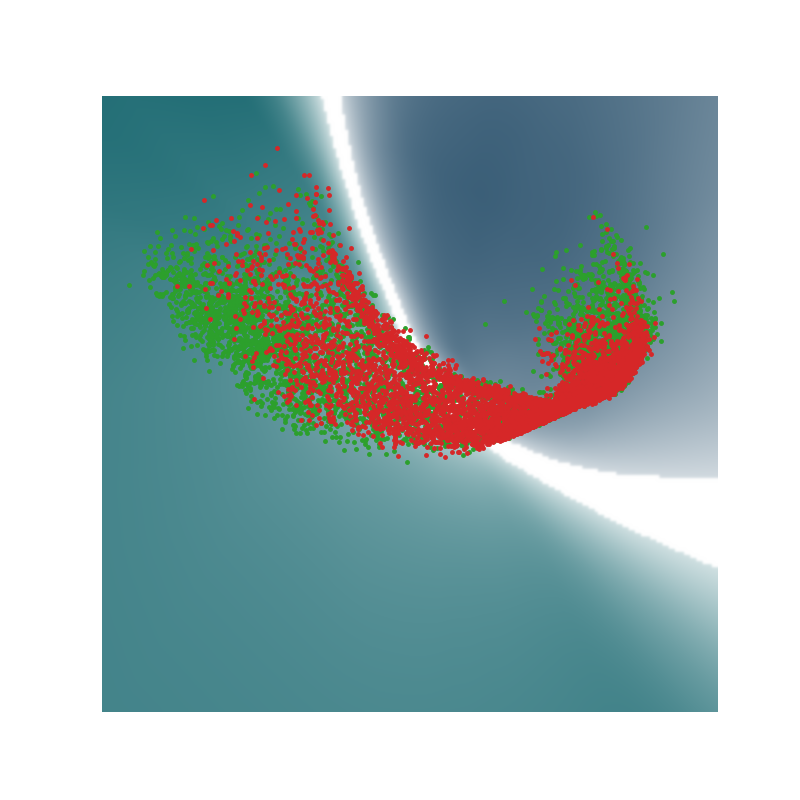}
         \caption{Context 2}
         \label{fig:tt_cr_2}
     \end{subfigure}
     \begin{subfigure}[b]{0.15\textwidth}
         \centering
         \includegraphics[width=\textwidth]{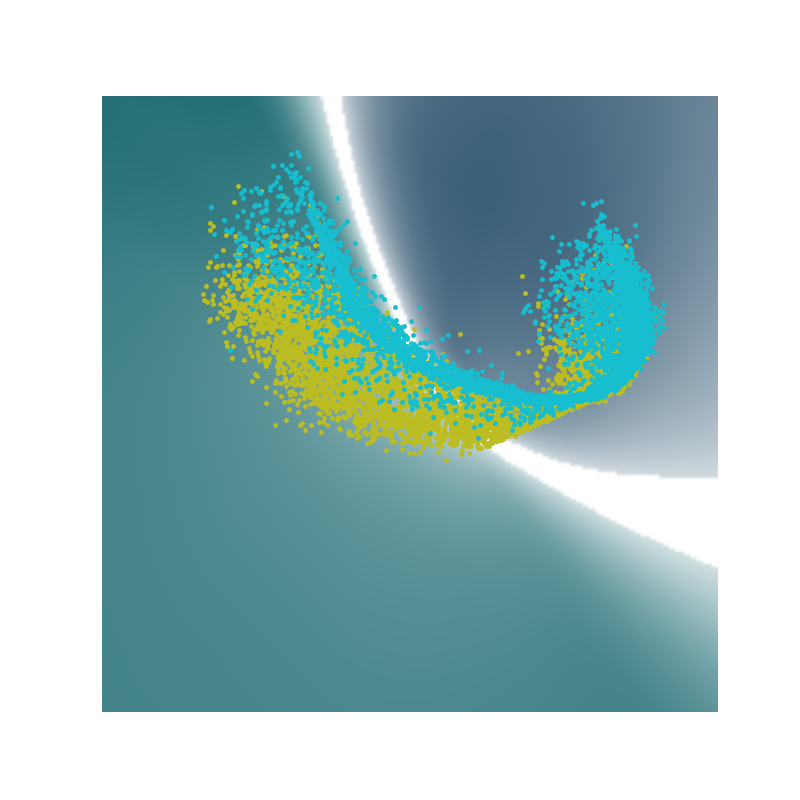}
         \caption{Context 3}
         \label{fig:tt_cr_5}
     \end{subfigure}
     \caption{Visualization of data from later contexts in the Class Incremental Learning setting using the ER strategy. The white regions represent the decision boundary, while the colored samples indicate data points from different classes.}
    \label{fig:tt-cr}
\end{figure}

\begin{figure}[h!]
     \centering
     \begin{subfigure}[b]{0.15\textwidth}
         \centering
         \includegraphics[width=\textwidth]{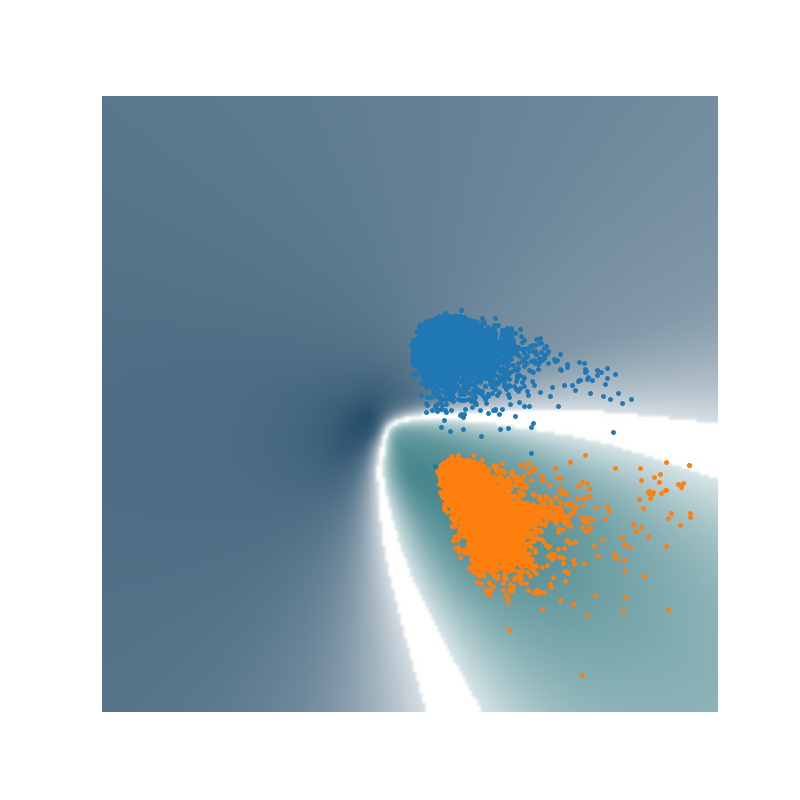}
         \caption{Context 1}
         \label{fig:tt_cf_1}
     \end{subfigure}
     \begin{subfigure}[b]{0.15\textwidth}
         \centering
         \includegraphics[width=\textwidth]{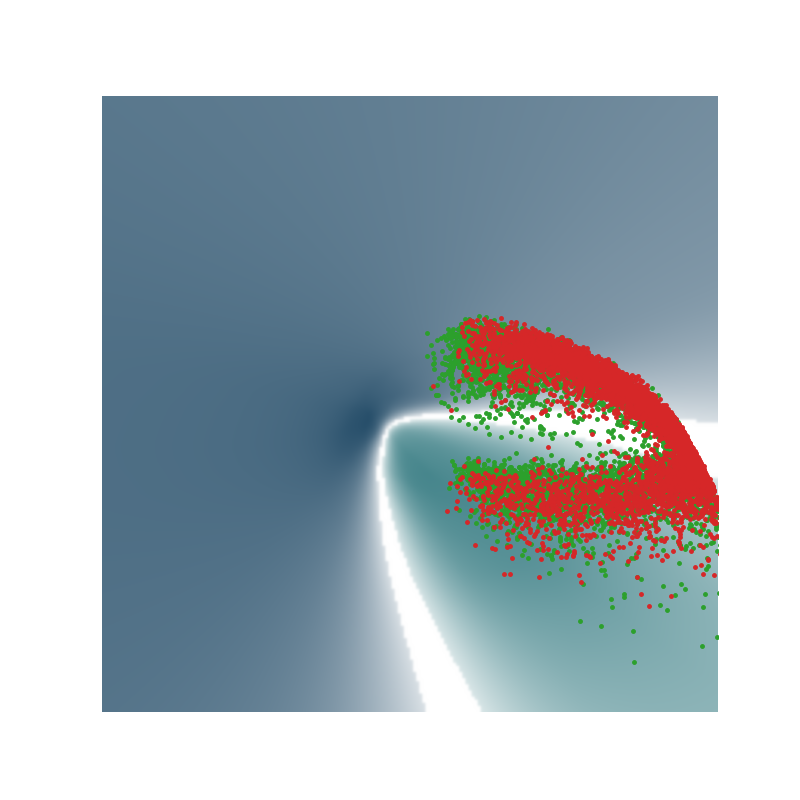}
         \caption{Context 2}
         \label{fig:tt_cf_2}
     \end{subfigure}
     \begin{subfigure}[b]{0.15\textwidth}
         \centering
         \includegraphics[width=\textwidth]{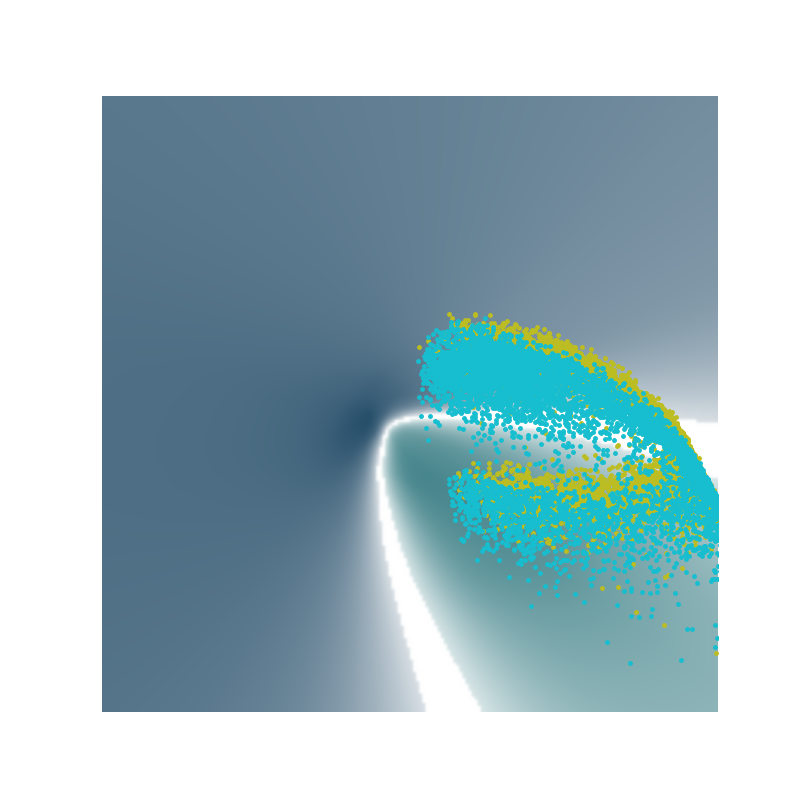}
         \caption{Context 3}
         \label{fig:tt_cf_5}
     \end{subfigure}
     \caption{Visualization of data from later contexts under the setting of Class incremental learning with strategy FROMP. The white regions represent the decision boundary, while the colored samples indicate data points from different classes.}
    \label{fig:tt-cf}
\end{figure}

\begin{itemize}
    \item \textbf{Experience Replay (ER)}: Figure~\ref{fig:tt-cr} shows that ER produces a feature space where class clusters are reasonably well-separated. The noticeable but controlled overlap between clusters suggests a plastic representation that can generalize and accommodate future tasks.
    \item \textbf{FROMP}: In contrast, Figure~\ref{fig:tt-cf} reveals that FROMP produces a feature space with severe overlap between class clusters. This lack of separability indicates poor plasticity, which may hinder the model's ability to distinguish new classes from old ones, correlating with its lower average accuracy in Table~\ref{tb:cl-accu}.
\end{itemize}

\tool's visualizations make the trade-offs between these CL strategies clear. ER maintains better plasticity at the cost of some stability, while FROMP's strong regularization leads to a less adaptable feature space, providing a visual explanation for their differing numerical performances.

\section{Proofs and Technical Details}
\label{app:proofs}

\subsection{Proof of Theorem~\ref{thm:reach} (homotopy preservation under $\varepsilon$-density)}
\label{app:reach-proof}
\textbf{Idea.} Consider the offset $\mathcal{M}_\alpha=\{x:\mathrm{dist}(x,\mathcal{M})\le \alpha\}$ for $\alpha<\tau$; the nearest-point projection onto $\mathcal{M}$ is well-defined and smooth. If $\varepsilon< c\tau$ with a small enough universal $c$, then the union of balls $\bigcup_{s\in S} B(s,\alpha)$ with $\alpha\in[2\varepsilon, \sqrt{\tau^2-\varepsilon^2}-\varepsilon]$ forms a good cover of $\mathcal{M}_\alpha$. By the Nerve Lemma \citep{Hatcher2002}, the nerve (\v{C}ech complex) of this cover is homotopy equivalent to $\mathcal{M}_\alpha$, which deformation retracts to $\mathcal{M}$. Standard interleaving bounds imply that a Vietoris--Rips complex at a commensurate scale shares the same homotopy type \citep{Hatcher2002,10.5555/3116660.3117013}. \qed

\subsection{Proof of Lemma~\ref{lem:eps-knn} (equivalence of $\varepsilon$ and $\overline{d}_k$)}
\label{app:eps-knn-proof}
\textbf{Upper bound.} Let $\lambda_{\min},\lambda_{\max}$ be lower/upper bounds on the sampling density on $\mathcal{M}$. A volumetric argument on geodesic balls shows that the typical $k$-th neighbor of a point in $S$ lies at radius $\asymp (k/n)^{1/d}$ (up to constants depending on curvature via reach and the density ratio $\lambda_{\max}/\lambda_{\min}$), hence $\overline{d}_k(S)\gtrsim \varepsilon(S)$ yields $C_1\,\overline{d}_k(S)\le \varepsilon(S)$ with high probability \citep{penrose1999k,penrose2003random}. \\
\textbf{Lower bound.} Conversely, an $\varepsilon$-net has at most constant overlap number depending on $d$; packing/covering duality gives that if $\varepsilon$ is small then each ball of radius $C\,\varepsilon$ contains $\Theta(k)$ samples whp, implying $\overline{d}_k(S)\lesssim \varepsilon(S)$. Combining the two directions gives $C_1\,\overline{d}_k(S)\le \varepsilon(S)\le C_2\,\overline{d}_k(S)$. \qed

\subsection{Proof of Corollary~\ref{cor:relden}}
The inequality \eqref{eq:eta-star} together with Lemma~\ref{lem:eps-knn} implies $\varepsilon(S)\le (C_2/C_1)\,\overline{d}_k(S) \le (C_2/C_1)\,\eta^\star\,\overline{d}_k(X) < c\,\tau$. Apply Theorem~\ref{thm:reach}. \qed

\subsection{Proof of Proposition~\ref{prop:phase} (phase transition)}
\label{app:phase-proof}
Let $n=rN$. By Lemma~\ref{lem:eps-knn}, $\varepsilon(S)$ increases continuously (in probability) as $n$ decreases, and there exists a unique $n_\mathrm{tip}$ (hence $r_\mathrm{tip}=n_\mathrm{tip}/N$) where $\varepsilon(S)=c\tau$. For $r>r_\mathrm{tip}$, Theorem~\ref{thm:reach} holds; for $r<r_\mathrm{tip}$ the condition fails and no homotopy window is guaranteed. Since Betti numbers of \v{C}ech/Rips complexes change only at discrete scale thresholds, the loss of the existence interval manifests as an abrupt change in observed topology and neighborhood preservation, i.e., a phase transition \citep{chazal2016structure}. \qed

\subsection{Binary search and monotonic feasibility}
\label{app:binary}
Define the feasible set $\mathcal{R}:=\{r\in(0,1]: \mathrm{RelDen}_k(S_r;X)\le \eta_{\mathrm{th}}\}$, where $S_r$ is a downsample of size $\lfloor rN\rfloor$.
Since $\overline{d}_k(S_r)$ is (stochastically) non-decreasing in $r\downarrow$, $\mathcal{R}$ is an interval $[r_\star,1]$. Thus, a standard binary search in $r$ returns the maximal feasible ratio $r_\star$ up to any prescribed tolerance.

\subsection{Choice of metric and cosine distance}
\label{app:metric}
If features are $\ell_2$-normalized, cosine distance equals the chordal distance on the unit sphere $\mathbb{S}^{h-1}$, which is bi-Lipschitz equivalent to the geodesic distance on sufficiently small neighborhoods. Therefore, $k$-NN radii and covering arguments used above transfer with modified constants, leaving the statements intact.

\paragraph{Choice of $k$ and metric.}
Take $k\gtrsim c\log n$ to ensure connectivity of the $k$-NN graph with high probability; small constants (e.g., $k\in[8,32]$) work well. With $\ell_2$-normalized features, cosine distance equals chordal distance on the sphere and is locally bi-Lipschitz to Euclidean, leaving the guarantees intact \citep{lee2006riemannian}.

% \subsection{Connectivity and the choice of $k$}
% For $n$ samples drawn from a regular density on a $d$-manifold, the $k$-NN graph is connected with high probability once $k\gtrsim c\log n$ for a universal constant $c$. This ensures stable neighborhood graphs in the pre-tipping regime and supports the use of small fixed $k$ (e.g., $k\in[8,32]$) in practice.

\paragraph{Limitations.}
If the sampling density is highly anisotropic (e.g., regions of much smaller local feature size than the global reach), a single global threshold $\eta_{\mathrm{th}}$ can be conservative. Region-wise density control can tighten the guarantee but is beyond our scope.

% \section*{Ethical Statement}
% There are no ethical issues.
% \section*{Acknowledgments}

\end{document}